%% file: main.tex
\documentclass[11pt]{article}
\input{header/packages}
\input{header/custom_defs}
\input{title/title}
\begin{document}
\maketitle
\input{title/abstract}
\pagenumbering{arabic} 
\input{paper/introduction}
\input{paper/prelim}
\input{paper/algorithm}
\input{paper/proof-overview}

\input{paper/expected-leave-count}

\input{paper/approx-factor}

\input{paper/lower-bound}
\input{paper/lb-ExKMC}

\bibliographystyle{plainnat}
\bibliography{references}
\end{document}

%% file: header/packages.tex
\usepackage[utf8]{inputenc}

\usepackage{fullpage}
\usepackage{mathtools}
\usepackage[margin = 2.5cm]{geometry}

\usepackage{amsmath, amssymb, amsthm}

\usepackage{graphicx}
\usepackage{color}
\usepackage[dvipsnames]{xcolor}

\usepackage{pgfplots}
\pgfplotsset{compat=1.15}

\usepackage{tikz}
\usepackage[hyperindex,breaklinks]{hyperref}

\usepackage[round]{natbib}

\usepackage{mdframed}

\usepackage{nicefrac}

\usepackage[shortlabels]{enumitem}

%% file: header/custom_defs.tex
\newtheorem{theorem}{Theorem}[section]
\newtheorem{claim}[theorem]{Claim}
\newtheorem{corollary}[theorem]{Corollary}
\newtheorem{lemma}[theorem]{Lemma}
\newtheorem{definition}[theorem]{Definition}

\newcommand{\pr}{\mathbb{P}}
\newcommand{\prob}[1]{\pr\left\{#1\right\}}
\newcommand{\E}{\mathbf{E}}

\newcommand{\one}{\mathbf{1}}

\newcommand{\bbR}{\mathbb{R}}

\DeclarePairedDelimiter\floor{\lfloor}{\rfloor}
\DeclarePairedDelimiter\ceil{\lceil}{\rceil}

\def\norm#1{\lVert#1\rVert}
\newcommand{\abs}[1]{\left\lvert#1\right\rvert}

\newcommand{\calE}{\mathcal{E}}

\newcommand{\calT}{\mathcal{T}}

\DeclareMathOperator{\cost}{cost}
\DeclareMathOperator{\OPT}{OPT}
\DeclareMathOperator{\sign}{sign}

\newcommand{\tempAlgName}{}
\newenvironment{myAlgorithm}[2]
{
\renewcommand{\tempAlgName}{#1}
\newcommand{\templabel}{#2}
\begin{figure}[ht]
\centering
\begin{mdframed}
\medskip
\noindent{\textbf{\tempAlgName}}}
{\smallskip
\end{mdframed}
\caption{\tempAlgName}
\label{\templabel}
\end{figure}}

\newcommand{\algInput}[1]{ \smallskip \noindent \textbf{Input: }#1\\}
\newcommand{\algOutput}[1]{\noindent \textbf{Output: }#1}

\newcommand{\Avg}{\operatorname{Avg}}

\tikzset{My Style/.style={shape=rectangle, rounded corners, align=center, draw, top color=white, bottom color=gray!20}}

%% file: title/title.tex
\title{Explainable $k$-means. Don't be greedy, plant bigger trees!}
\author{Konstantin Makarychev\footnote{Equal contribution. The authors were supported by NSF Awards CCF-1955351 and CCF-1934931.} \and Liren Shan\footnotemark[1]}
\date{Northwestern University\\Evanston, IL}

%% file: title/abstract.tex
\begin{abstract}
We provide a new \emph{bi-criteria} $\tilde{O}(\log^2 k)$ competitive algorithm for explainable $k$-means clustering. Explainable $k$-means was recently introduced by Dasgupta, Frost, Moshkovitz, and Rashtchian (ICML 2020). It is described by an easy to interpret and understand (threshold) decision tree or diagram. The cost of the \emph{explainable $k$-means} clustering equals to the sum of costs of its clusters; and the cost of each cluster equals the sum of squared distances from the points in the cluster to the center of that cluster. The best non bi-criteria algorithm for explainable clustering $\tilde{O}(k)$ competitive, and this bound is tight.

Our randomized bi-criteria algorithm constructs a threshold decision tree that partitions the data set into $(1+\delta)k$ clusters (where $\delta\in (0,1)$ is a parameter of the algorithm). The cost of this  clustering is at most $\tilde{O}(\nicefrac{1}{\delta}\cdot \log^2 k)$ times the cost of the optimal unconstrained $k$-means clustering. We show that this bound is almost optimal.
\end{abstract}

%% file: paper/introduction.tex
\section{Introduction}
In this paper, we study explainable $k$-means clustering. $k$-means is one of the most popular ways to cluster data. It is widely used in data science and machine learning. A $k$-means clustering of data set $X$ in $\bbR^d$ is determined by its $k$ centers $c^1,c^2,\dots, c^k$. Specifically, $k$-means clustering is a $d$-dimensional Voronoi diagram for centers $c^1,c^2,\dots, c^k$, in which,  the $i$-th cluster contains those points in $X$ that are closer to $c^i$ than to any other center $c^j$. The cost of the clustering equals 
\begin{equation}\label{eq:cost-k-means}
\cost(X; c^1,c^2,\dots, c^k) \equiv \sum_{i=1}^d\sum_{x\in P_i} \|x - c^i\|_2^2,
\end{equation}
where $P_i$ is the $i$-th cluster.

In a recent ICML paper, \cite*{DFMR20} observed that it can be hard for a human to understand $k$-means clustering. Clusters in $k$-means are determined by all features (coordinates) of the data. Thus, usually there is no a concise explanation of why a particular point belongs to one cluster or another. To make $k$-means more understandable for humans, \cite{DFMR20} proposed an alternative way to cluster data, which they called \emph{explainable $k$-means}. In \emph{explainable $k$-means}, the data set is partitioned into clusters using a threshold decision tree with $k$ leaves (a variant of a binary space partitioning tree). Every internal node $u$ of the tree splits the data into two disjoint groups based on a single feature (coordinate). A point $x$ is assigned to the left subtree of $u$, if $x_i \leq \theta$; it is assigned to the right subtree of $u$, if $x_i > \theta$. Points assigned to each of the $k$ leaves form a cluster. The cost of explainable $k$-means clustering is defined in the same way as for $k$-means. It is equal to the sum of cluster costs:
$$\cost(X, \calT) = \sum_{i=1}^d\sum_{x\in P_i} \|x - c^i\|_2^2,$$
where $P_1,\dots,P_k$ are clusters; $c^1,\dots, c^k$ are centers of $P_1,\dots, P_k$; and $\calT$ is the decision tree that defines the clustering. 

Note that explainable $k$-means clustering can be represented by a simple decision diagram as in Figure~\ref{fig:diagram-example}. This diagram is easy to understand, and humans can easily determine to which cluster a given data point $x$ belongs to.

\input{figure/example}
\input{figure/kmeansplusplus}

The cost of explainability or the competitive ratio of an explainable $k$-means clustering is the ratio between the cost of that clustering and the cost of the optimal unconstrained $k$-means clustering for the same data set.
\cite{DFMR20} showed how to obtain a $k$-means clustering with a competitive ratio of $O(k^2)$. This competitive ratio was improved to a near-optimal\footnote{It is possible to get a better competitive ratio for low dimensional data. For details, see Section~\ref{sec:related-work}}
bound of~$\tilde{O}(k)$ by \cite*{MS21}; \cite*{GJPS21}; and \cite*{EMN22}. This guarantee does not depend on the size and dimension of the data set.  However, it is large for large data sets. For comparison, the best competitive ratio for explainable \emph{$k$-medians} is exponentially better than $\tilde O(k)$. It equals $\tilde{O}(\log k)$ (see \cite{MS21,EMN22}). Nevertheless, \cite{DFMR20} and then \cite{frost2020exkmc} empirically demonstrated that, in practice, the price of explainability for $k$-means clustering is fairly small. In this work, we provide a theoretical justification for this observation. Specifically, we show a bi-criteria approximation algorithm which finds a decision tree with $(1+\delta)k$ leaves and has a competitive ratio of $O(\nicefrac{1}{\delta}\log^2 k \log\log k)$, where $\delta$ is a parameter between $0$ and $1$.

We note that in practice the cost of the optimal $k$-means clustering is approximately the same for $k$ and $(1+\delta)k$ clusters (here $\delta\in (0,1)$ is a small constant). In other words, for many data sets~$X$, we have 
$\OPT_k(X)\approx \OPT_{(1+\delta)k}(X)$, where $\OPT_k(X)$ is the cost of the optimal unconstrained $k$-means clustering of $X$ with $k$ clusters\footnote{In the worst case, we may have $\OPT_{(1+\delta)k}(X)\ll \OPT_k(X)$. For example, if $X$ contains exactly $(1+\delta)k$ points, then $\OPT_{(1+\delta)k}(X) = 0$ but $\OPT_{k}(X) > 0$.}. The plot in Figure~\ref{fig:kmeansplusplus} shows that the cost of $k$-means++ clustering for BioTest data set from KDD Cup  \citep{kddcup2004} is about the same for $k$ and $(1+\delta)k$ centers when $k$ is between $50$ and $200$. If $\OPT_k(X)\approx \OPT_{(1+\delta)k}(X)$, then our algorithm gives a \emph{true}
$\tilde{O}(\log^2 k)$ approximation, because
$$\cost(X, \calT) \leq 
\tilde{O}(\log^2 k)
\OPT_{k}(X) \approx
\tilde{O}(\log^2 k)
\OPT_{(1+\delta)k}(X).$$

\subsection{Our Results}
We now formally state our results. We provide a randomized algorithm for finding bi-criteria explainable $k$-means. Similarly to the algorithm by \cite{frost2020exkmc},  our  algorithm takes $k$ centers $\{c^1,c^2,\dots,c^k\}$ and a parameter $\delta > 0$ and returns a threshold decision tree with $(1+\delta)k$ leaves. Each leaf of the tree is labeled with one of the centers $c^1,c^2,\dots,c^k$. Let us denote the center returned by the 
decision tree $\calT$ for point $x$ by $\calT(x)$. Then, the cost of explainable clustering defined by $\calT$ equals 
\begin{equation}\label{eq:cost-threshold-tree}
\cost(X,\calT)\equiv \sum_{x\in X} \|x-\calT(x)\|_2^2.
\end{equation}
\begin{theorem}
There exists a polynomial-time randomized algorithm that given a data set $X$, a set of $k$ centers 
$C = \{c^1,c^2,\dots,c^k\}$, and parameter $\delta \in (0,1)$, creates a threshold decision tree $\calT$ whose leaves are labeled with centers from $C$. The expected number of leaves in $\calT$ is $(1+\delta)k$, and the expected cost of explainable clustering defined by $\calT$ is
\begin{equation*}
\E[\cost(X,\calT)] \leq O(\nicefrac{1}{\delta} \cdot \log^2 k \log\log k) \cdot \cost(X,C).
\end{equation*}
\end{theorem}
Observe that our algorithm constructs a tree with $(1+\delta)k$ leaves and only $k$ centers. Thus, we can use this algorithm to partition $X$ into $k$ clusters. In this case, one cluster may be assigned to several different leaves. Alternatively, we can assign its own cluster to every leaf. Then, we will have a proper threshold decision tree with $(1+\delta)k$ clusters.
In either case, we can further improve the clustering by replacing the original center $c^i$ assigned to each leaf $u$ with the optimal center for the cluster assigned to $u$ (the optimal center is the centroid of that cluster).

If $C$ is the optimal set of centers for $k$ means, then the explainable clustering provided by our algorithm has an expected cost of at most 
$O(\nicefrac{1}{\delta} \cdot \log^2 k \log\log k)\OPT_k(X)$. Furthermore, 
if $C$ is obtained by a constant factor bi-criteria approximation algorithm such as $k$-means++ (in which case, $|C|=(1+\delta)k$ and $\cost(X,C)\leq O(1)\cdot \OPT_k(X)$), then the expected
cost of the explainable clustering is also at most $O(\nicefrac{1}{\delta} \cdot \log^2 k \log\log k)\OPT_k(X)$ and the number of leaves in the threshold decision tree is at most $(1+3\delta)k$ in expectation.
 
As we note above, our work is influenced by the paper of \cite*{frost2020exkmc}, who showed a bi-criteria algorithm for explainable $k$-means. However, our algorithm for this problem is very different from theirs. It uses the approach from our previous paper (\citet*{MS21}). In that paper, we gave
an algorithm for finding explainable \emph{$k$-medians with $\ell_2$ norm}. Our new algorithm has an additional crucial step: It duplicates some centers when the algorithm splits nodes. This step gives an exponential improvement to the competitive ratio for $k$-means. The analysis of our algorithm is considerably more involved than the analysis of the previous algorithm.

We complement our algorithmic results with an almost matching lower bound of $\Omega(\nicefrac{1}{\delta} \cdot \log^2 k)$ for all threshold trees with at most $(1+\delta)k$ leaves.

\begin{theorem}\label{thm:lb_bi_criteria}
For every $k > 500$ and $\ln^3 k / \sqrt{k} < \delta < 1 / 100$, there exists an instance $X$ with $k$ clusters such that the $k$-means cost for every threshold tree $\calT$ with $(1 + \delta)k$ leaves is at least
$$
\mathrm{cost}(X,\calT) \geq \Omega\left(\frac{\log^2 k}{\delta}\right)\OPT_k(X).
$$
\end{theorem}

In Section~\ref{sec:greedy-lb}, we provide a family of $k$-means instances for which a greedy bi-criteria algorithm finds a solution of cost 
$\mathrm{cost}(X,\calT) \geq \tilde{\Omega}(k^{2})\OPT_k(X)
$ for $k\to \infty$.

\subsection{Related Work}\label{sec:related-work}
Decision trees have been widely used for classification and clustering due to their simplicity. Examples of decision tree algorithms for supervised classification include CART by~\cite{breiman-classification}, ID3 by~\cite{quinlan1986induction}, and C4.5 by~\cite{quinlan1993c4}. Examples of decision tree algorithms for unsupervised clustering include algorithms by~\cite{liu2005clustering}, \cite{fraiman2013interpretable}, Silhouette Metric~(\citet{bertsimas2018interpretable}), \cite{saisubramanian2020balancing}.   

\cite{DFMR20} proposed the problems of explainable $k$-means and $k$-medians clustering in $\ell_1$. They defined these problems and offered algorithms for explainable $k$-means and $k$-medians with the competitive ratios of $O(k^2)$ and $O(k)$, respectively. Later, \cite{frost2020exkmc} designed a new bi-criteria algorithm for these problems and evaluated its performance in practice. \cite*{laber2021price}, \cite*{MS21}, \cite*{CharikarHu21}, \cite*{EMN22}, and \cite*{GJPS21} provided improved upper and lower bounds for explainable $k$-means and $k$-medians. The best competitive ratios for explainable  $k$-means and $k$-medians are $\tilde O(k)$ and $\tilde O(\log k)$, respectively. \cite{MS21},  \cite{EMN22}, and \cite{GJPS21} gave a $\tilde{O}(k)$ competitive ratio for explainable $k$-means; and \cite{MS21} and  \cite{EMN22} gave a $\tilde{O}(\log k)$ bound for $k$-medians. \cite*{CharikarHu21} provided
$k^{1-\nicefrac{2}{d}}\cdot \operatorname{poly}(d \log k)$ algorithm for $k$-means (this algorithm gives stronger approximation guarantees when the dimension of the space, $d$, is small). Additionally, \cite{MS21} gave an $\tilde O(\log^{\nicefrac{3}{2}} n)$ competitive algorithm for explainable $k$-medians in $\ell_2$.

\citet{bmd09}, \citet{BZMD14}, \citet{CEMMP15}, \citet{MMR19} and \citet{BBCGS19} showed how to reduce the dimensionality of a data set for $k$-means clustering. Particularly, \citet{MMR19} proved that we can use the Johnson--Lindenstrauss transform to reduce the dimensionality of $k$-means to $d'= O(\log k)$. Note, however, that the Johnson--Lindenstrauss transform cannot be used for the explainable $k$-means, because this transform does not preserve the set of features. Instead, one can use a \emph{feature selection} algorithm by~\citet{BZMD14} or \citet{CEMMP15} to reduce the dimensionality to 
$d' = \tilde{O}(k)$.

The classic $k$-means clustering has been extensively studied by researchers in machine learning and theoretical computer science. Lloyd's algorithm (\cite{lloyd1982least}) is the most popular heuristic  for $k$-means clustering. \cite{arthur2007k} proposed a randomized seeding algorithm called $k$-means++, which achieves an expected $O(\log k)$ approximation. \cite*{ahmadian2019better} designed a primal-dual algorithm with an approximation factor of $6.357$. It was recently improved to $6.12903$ by \cite*{GORSV22}. \cite{dasgupta2008hardness} and \cite*{aloise2009np} showed that $k$-means problem is NP-hard. \cite{awasthi2015hardness} showed that it is also NP-hard to approximate the $k$-means objective within a factor of $(1+\varepsilon)$ for some positive constant $\varepsilon$ (see also~\cite*{lee2017improved}). The bi-criteria approximation for $k$-means has also been studied before. \cite*{aggarwal2009adaptive} proved  that $k$-means++ that picks $(1+\delta)k$ centers gives a constant factor bi-criteria approximation for some constant $\delta > 0$. Later, \cite{wei2016constant} and \cite*{MRS21} gave improved bi-criteria approximation guarantees for $k$-means++. \cite*{MMSW16} designed local search and LP-based algorithms with better bi-criteria approximation guarantees.

%% file: figure/example.tex
\begin{figure}\label{fig:diagram-example}
    \centering
    \includegraphics[width=0.35\linewidth]{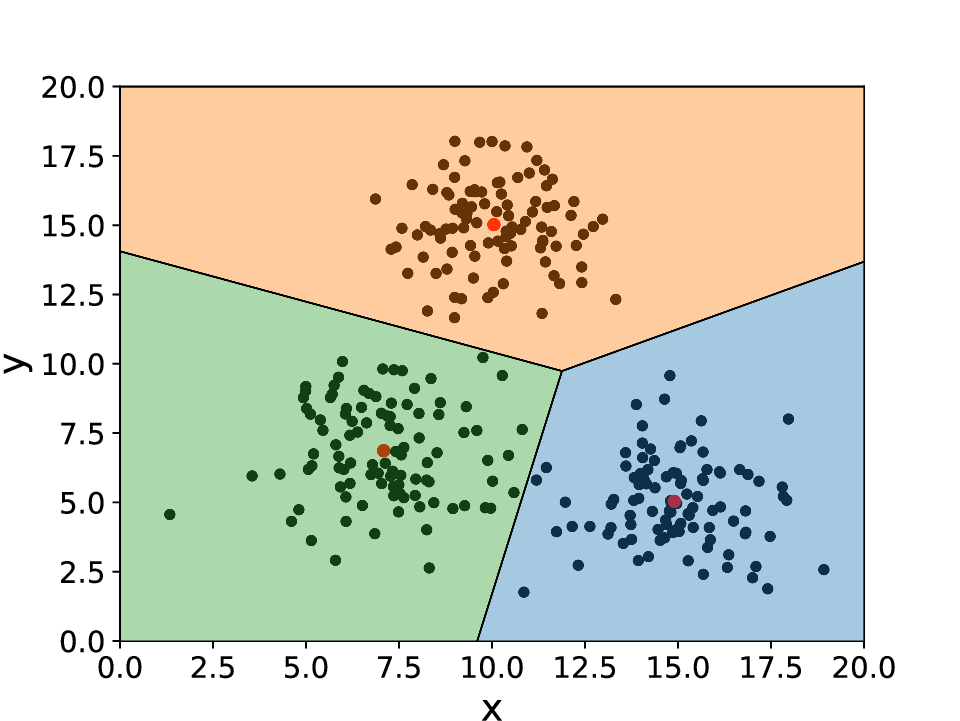}
    \includegraphics[width=0.35\linewidth]{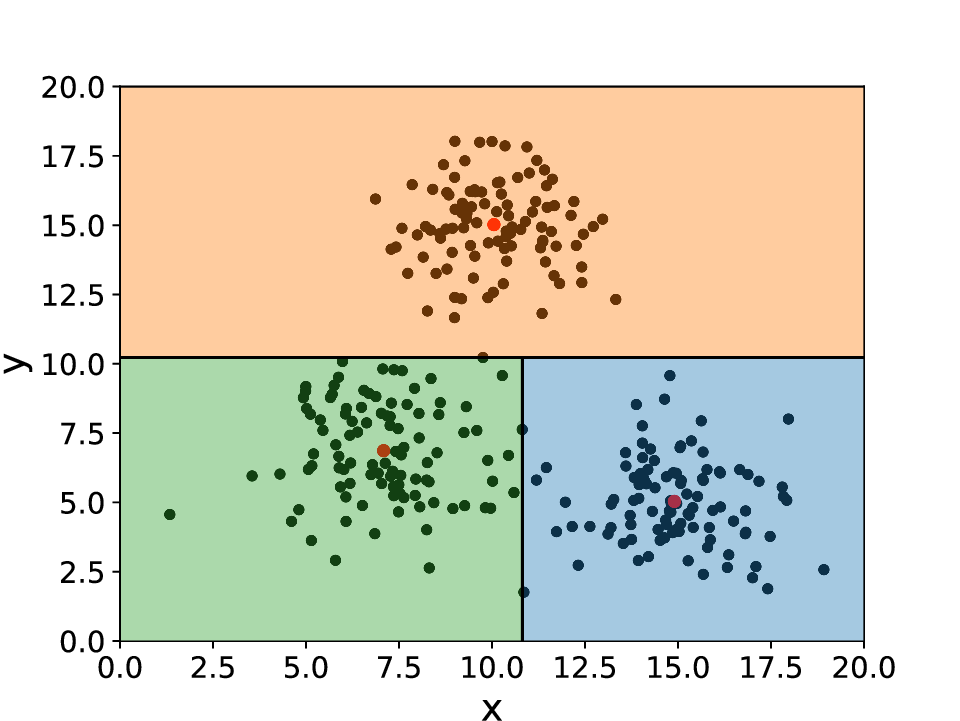}
    \begin{tikzpicture}
    \node[My Style]{$y\leq 10.2$}
    child{node[My Style]{$x \leq 10.8$}
    child{node[style={shape=rectangle, rounded corners, align=center, draw, fill=green!20}]{$2$}}
    child{node[style={shape=rectangle, rounded corners, align=center, draw, fill=blue!20}]{$3$}}}
    child{node[style={shape=rectangle, rounded corners, align=center, draw, fill=orange!20}]{$1$}};
    \end{tikzpicture}
    \caption{Explainable and non-explainable $k$-means. The left diagram shows the optimal Voronoi partition of the plane. The middle diagram shows an explainable partition. The right diagram shows the corresponding decision tree for explainable clustering.}
    \label{fig:my_label}
\end{figure}
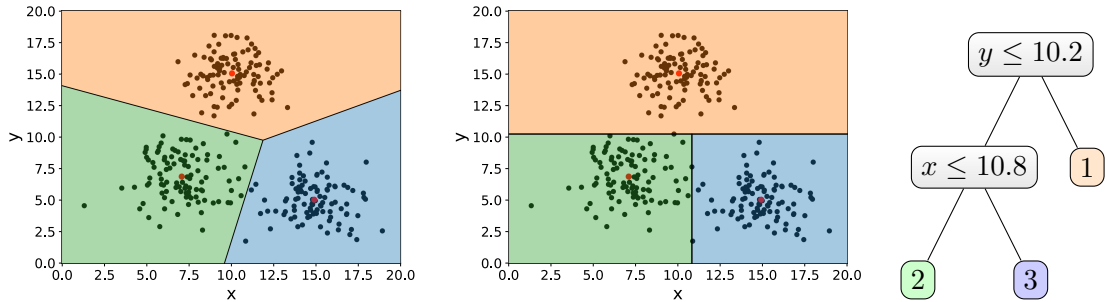

%% file: figure/kmeansplusplus.tex
\begin{figure}
    \centering
    \begin{minipage}{0.45\linewidth}
    \begin{tikzpicture}[scale=0.7]
        \begin{axis}[
        title= {BioTest},
        xlabel={\#centers},
        ylabel={cost},
        grid = major]
        \addplot[black,domain=1:2, line width = 1pt]  table[x=centers,y=avgKMeansPP,col sep=comma] {figure/bio-test-results.csv}; \addlegendentry{$k$-means++}
        \end{axis}
    \end{tikzpicture}
    \end{minipage}
    \quad
    \begin{minipage}{0.45\linewidth}
    \begin{tikzpicture}[scale=0.7]
        \begin{axis}[
        title= {BioTest},
        xlabel={\#centers},
        ylabel={ratio},
        grid = major]
        \addplot[black,domain=1:2, line width = 1pt]  table[x=centers,y=ratios,col sep=comma] {figure/bio-test-ratio2.csv}; \addlegendentry{$k$-means++}
        \end{axis}
    \end{tikzpicture}
    \end{minipage}
    \caption{Performance of $k$-means++ on BioTest data set. The left diagram shows the cost of $k$-means++ for $k= 5,10,15,\dots,200$. The clustering cost is divided by the cost of $k$-means with 1000 clusters. The right diagram shows the ratio between the clustering cost with $k$ centers and the cost with $(1+\delta)k$ centers for $k=5,10,\dots, 150$ and $\delta = 0.2$. }
    \label{fig:kmeansplusplus}
\end{figure}
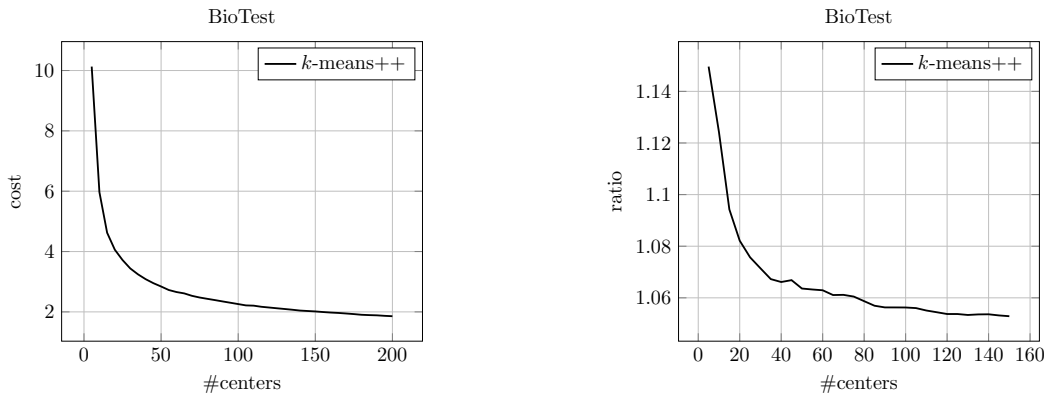

%% file: paper/prelim.tex
\section{Preliminaries}\label{sec:prelim}
Consider a set of points $X\subseteq \bbR^d$ and an integer $k>1$. A $k$-means clustering consists of $k$ clusters $P_1,\dots, P_k$. Each cluster $P_i$ is assigned a center $c^i$, which is the centroid (geometric center) of $P_i$. The cost of the clustering equals
$$
\cost(X; c^1,\dots,c^k) \equiv \sum_{i=1}^d\sum_{x\in P_i} \|x - c^i\|_2^2.$$
The optimal $k$-means clustering is the clustering of the minimum cost. We denote the cost of the optimal $k$-means clustering with $k$ clusters by $\OPT_k(X)$.

A threshold decision tree is a tree that recursively partitions $\bbR^d$ into cells using hyperplane cuts. Every node in the tree corresponds to a cell (polytope) of the space. The root corresponds to the entire space $\bbR^d$. In this paper, we will identify nodes of the tree with the  cells they correspond to. Thus, a threshold decision tree defines a hierarchical partitioning of $\bbR^d$ into $k$ cells or clusters. 

Each internal node (cell) $u$ in the threshold tree is split into two nodes 
$u_{left}$ and $u_{right}$
using a threshold cut $(i, \xi)$ as follows:
$$u_{left} = \{x \in u: x_i \leq \xi\} 
\;\;\;\text{ and }\;\;\;
u_{right} = \{x \in u: x_i > \xi\}.$$
We assign a center $c$ to every leaf of the threshold decision tree. Let $\calT(x)$ (where $x\in \bbR^d$) be the center assigned to the unique leaf $u$ of $\calT$ that contains $x$. In this paper, we will also assign centers to internal nodes of the tree. We will denote the set of centers assigned to node $u$ by $C_u$. For leaf nodes, we have $|C_u|=1$. 

Consider a data set $X$ and threshold decision tree $\calT$. The $k$-means cost of $\calT$ equals
$$
\cost(X,\calT)\equiv \sum_{x\in X} \|x-\calT(x)\|_2^2.$$
The competitive ratio of explainable clustering defined by $\calT$
is $\cost(X,\calT)/\OPT_k(X)$. We say that a randomized algorithm is $\alpha$-competitive if the expected cost of the explainable clustering returned by the algorithm is at most $\alpha \cost(X,C)$, where $C$ is the set of centers provided to the algorithm. 

A bi-criteria solution to explainable $k$-means clustering with parameter $\delta$ is a threshold decision tree with at most $(1+\delta)k$ leaves. In this paper, we describe an algorithm that finds a tree with at most $(1+\delta)k$ leaves and $k$ distinct centers assigned to them.

%% file: paper/algorithm.tex
\section{Algorithm}\label{sec:alg}

In this section, we present an algorithm for explainable $k$-means clustering. The input of the algorithm is a set of centers $c^1,\dots,c^k$ and parameter $\delta$. The output is a threshold decision tree in which every leaf node is labeled with one of the centers $c^i$. In Sections~\ref{sec:exp-number-leaves} and~\ref{sec:approx-factor}, we will show that the expected number of leaves in the decision tree is $(1+\delta)k$ and the approximation factor of the obtained clustering is $O(\nicefrac{1}{\delta}\cdot \log^2 k\cdot \log \log k)$.

\medskip

\noindent\textbf{Algorithm.} Our algorithm builds a binary threshold tree using a top-down approach. The algorithm assigns every node $u$ in the tree a subset of centers $c^1,\dots,c^k$. We denote this subset $C_u$. First, the algorithm creates a tree $\calT_1$ with a root vertex $r$ and assigns all centers $c^1, c^2, \dots, c^k$ to it. Then, the algorithm recursively splits leaf nodes in the threshold tree until each leaf is assigned exactly one center. At each step $t$, the algorithm chooses a coordinate $i_t \in \{1,2,\dots, d\}$, a positive threshold $\theta_t \in (0,1)$, and number $\sigma_t$ in $\{\pm 1\}$ uniformly at random. For each leaf $u$ with more than one center, it calls function \emph{Divide-and-Share} to split node $u$ into two parts. 

\input{algorithms/build-tree}
\input{algorithms/divide-and-share}

Function \emph{Divide-and-Share} first finds a median\footnote{Median $m^u$ satisfies the following property: For ever coordinate $i$, each of the sets $\{c\in C_u: c_i < m^u_i\}$ and $\{c\in C_u: c_i > m^u_i\}$ contains at most half of all points from $C_u$.} of all centers assigned to $u$, which we denote by $m^u$. Let $R_u$ be the maximum distance from centers in node $u$ to the median $m^u$. The algorithm creates two child nodes for $u$ using cut $\omega_t=(i_t, \xi_t)$ with $\xi_t = m^u_i + \sigma_t \sqrt{\theta_t} R_u$. Then, \emph{Divide-and-Share} assigns two sets of centers, $Left$ and $Right$, defined in Figure~\ref{alg:divide-and-share} to the left and right children of $u$, respectively.
Note that these sets share centers in the strip of width $2\varepsilon \sqrt{\theta_t} R_u$:
$$Left\cap Right = \{c\in C_u:
(m^u_i  + \sigma_t \sqrt{\theta_t} R_u) - \varepsilon \sqrt{\theta_t} R_u \leq
c_i \leq
(m^u_i  + \sigma_t \sqrt{\theta_t} R_u) + \varepsilon \sqrt{\theta_t} R_u
\}.
$$
If one of the sets, $Left$ or $Right$, is empty, then \emph{Divide-and-Share} discards both newly created children of $u$.

We show that the bi-criteria approximation factor of the algorithm is $O(\nicefrac{1}{\delta}\log^2 k\log\log k)$ and the expected number of leaves is $(1 + \delta)k$. In the next section, we give a proof overview. Then, we prove the upper bounds on the expected number of leaves and approximation factor of the algorithm in Sections~\ref{sec:exp-number-leaves} and~\ref{sec:approx-factor}, respectively.

%% file: algorithms/build-tree.tex
\begin{myAlgorithm}{Threshold Tree Construction}{alg:build-tree}

\algInput{a data set $X$ and set of centers $C = \{c^1,c^2,\dots,c^k\}$, a parameter $\delta \in (0,1)$}
\algOutput{a threshold tree $\calT$}
\begin{itemize}
    \item Create a tree $\calT_1$ containing a root $r$. Let $C_r = C$.
    \item \textbf{while} $\calT_t$ contains a leaf with at least two distint centers \textbf{do}:  
    \begin{itemize}
       \item Sample $i_t \in \{1,2,\dots,d\}$, $\theta_t \in (0,1)$, and $\sigma_t\in\{\pm 1\}$ uniformly at random.
       \item For each leaf $u$ in the tree $\calT_t$ containing more than one center, split node $u$ using \emph{Divide-and-Share} with parameters $u$, $i_t$, $\theta_t$, $\sigma_t$, and $\varepsilon = \min\{\nicefrac{\delta}{15\ln k},\nicefrac{1}{384}\}$.
       \item Update $t= t+1$.
    \end{itemize}
\end{itemize}
\end{myAlgorithm}

%% file: algorithms/divide-and-share.tex
\begin{myAlgorithm}{Function Divide-and-Share}{alg:divide-and-share}

\algInput{a node $u$, a coordinate $i\in\{1,\dots,d\}$, a positive threshold $\theta$, a number $\sigma\in\{\pm 1\}$, and a parameter $\varepsilon$}
\algOutput{if successful, the function splits $u$ into two parts}
\begin{itemize}
\item Find the median of all centers assigned to node $u$. Denote it by $m^u$.
\item Let $R_u = \max\{\|c-m^u\|_2: c \in C_u\}$ be the maximum distance from $m^u$ to one of the centers in $C_u$.
\item Let
\begin{align*}
    Left&=\{c\in C_u: c_i \leq m^u_i  + \sigma \sqrt{\theta} R_u + \varepsilon \sqrt{\theta} R_u \};\\
    Right&=\{c\in C_u: c_i \geq m^u_i + \sigma \sqrt{\theta} R_u -\varepsilon \sqrt{\theta} R_u \}.
\end{align*}
\item If both sets -- $Left$ and $Right$ -- are nonempty, then 
\begin{itemize}
    \item Split $u$ into two parts using cut $(i,m^u + \sigma \sqrt{\theta} R_u$).
    \item Assign the set of centers $Left$ to the left child $u_{left}$ and the set of centers $Right$ to the right child, $u_{right}$.
\end{itemize}
\item Otherwise, return the unmodified tree (in this case, we say that Divide-and-Share \emph{fails}).   
\end{itemize}
\end{myAlgorithm}

%% file: paper/proof-overview.tex
\section{Proof Overview}\label{sec:proof-overview}

In this section, we provide an overview of the analysis of our algorithm, give definitions, and discuss the motivation for the proofs. In Sections \ref{sec:exp-number-leaves} and \ref{sec:approx-factor}, we present detailed proofs.

\subsection{Cost of Clustering}
We first analyze approximation guarantees for our algorithm. We show that the expected approximation factor is $O(\nicefrac{1}{\delta}\log^2 k \log\log k) = O(\nicefrac{1}{\varepsilon}\log k \log\log k)$, particularly for constant $\delta$ (e.g., $\delta = 0.05$), the expected approximation factor is $O(\log^2 k \log\log k)$. We denote the final tree returned by the algorithm by $\calT$. Let $\calT(x)$ be the center assigned by the threshold tree $\calT$ to point $x$.
\begin{theorem}\label{thm:main-approx-factor}
For every set of centers $c^1,\dots,c^k$ in $\bbR^d$, every $\delta\in (0,1)$, and every $x\in \bbR^d$, we have
\begin{equation}\label{eq:thm-aprox-factor}
\E\Big[\|x-\calT(x)\|_2^2\Big]\leq 
O(\nicefrac{1}{\delta}\;\log^2 k \log\log k) \min_{c\in\{c^1,\dots,c^k\}} \|x - c\|_2^2.
\end{equation}
\end{theorem}
This theorem guarantees that the expected approximation factor for every point $x$ is at most $O(\nicefrac{1}{\delta}\;\log^2 k \log\log k)$. Consequently, the expected approximation factor for any data set $X$ is also bounded by $O(\nicefrac{1}{\delta}\;\log^2 k \log\log k)$.

Fix an arbitrary point $x$ for the entire proof of Theorem~\ref{thm:main-approx-factor}. If $x$ equals one of the centers $c^i$, then $\calT(x)$ also always equals $c^i$. Hence, $\|x - \calT(x)\|_2^2 =0$ and bound (\ref{eq:thm-aprox-factor}) trivially holds. So, from now on, we will assume that $x$ is not one of the centers.

Denote by $\calT_t$ the tree built by the algorithm in the first $(t-1)$ steps. Tree $\calT_1$ contains only one node -- the root. The root corresponds to the entire space $\bbR^d$ and all centers $c^1,\dots,c^k$ are assigned to it. Since point $x$ is fixed, we will only consider nodes $u$ in $\calT$ that contain $x$. Let $u_t$ be the leaf node of the tree $\calT_t$ that contains $x$. That is, $u_t$ is the leaf node that contains $x$ at the beginning of iteration $t$. Nodes $u_1,u_2,\dots$ form a path in the tree $\calT$ from the root to the unique leaf of $\calT$ that contains $x$. To simplify notation, we denote 
$$C_t=C_{u_t},\;\;R_t = R_{u_t},\;\;m^t = m^{u_t}.$$ Also, let $D_t$ be the diameter of set $C_t$:
$$D_t=\max\{\|c'-c''\|_2: c',c''\in C_t\}.$$
Finally, let $\calT_t(x)$ be the closest center from the set $C_t$ to point $x$. We call this center the tentative center for point $x$ at step $t$. The tentative cost of $x$ at step $t$ is $\|x-\calT_t(x)\|_2^2$. 

Initially, at step $1$, the tentative center for point $x$ is the closest center $c\in\{c^1,\dots,c^k\}$ to $x$. If the tentative center for $x$ does not change, then the eventual cost of $x$, $\|x-\calT(x)\|_2^2$ exactly equals the optimal cost $\|x-c\|_2^2$. However, at some step $t$, point $x$ may be separated from its tentative center $c$ (see below for a formal definition), in which case another tentative center $\calT_{t+1}(x)$ is assigned to $x$. At this step, the tentative cost of $x$ may significantly increase. Moreover, the tentative cost of $x$ may further increase if $x$ is separated from the new tentative center. Our goal is to give an upper bound on the expected total cost increase.

\begin{definition}\label{def:separation}
We say that $x$ is separated from its tentative center $c=\calT_t(x)$ at step $t$, if $c\notin C_{t+1}$. 
\end{definition}
Note that $x$ is separated from its tentative center $c=\calT_t(x)$ at step $t$ if and only if $c$ is no longer the tentative center for $x$ at step $t+1$ ( $\calT_{t+1}(x)\neq \calT_t(x)$). We now define $A_k$. Loosely, speaking $A_k$ is the approximation factor of the algorithm for the given set of centers $c^1,\dots,c^k$ and point $x$. For technical reasons, the formal definition is more involved.

\begin{definition}\label{def:Ak}
Let $A_k$ be the smallest number such that the following inequality holds with probability $1$ for every partially built tree $\calT_t$:
\begin{equation}\label{eq:def:Ak}
\E\Big[\|x-\calT(x)\|_2^2 \mid \calT_t\Big]
\leq 
A_k \; \|x-\calT_t(x)\|_2^2.
\end{equation}
\end{definition}
In this definition, $\E\Big[\|x-\calT(x)\|_2^2 \mid \calT_t\Big]$ is the conditional expectation of the eventual cost of $x$ given that at step $t$ the partially built tree is $\calT_t$. Thus, if at some step $t$, the tentative center for $x$ is $c$, then the expected final cost $\E[\|x-\calT(x)\|_2^2\mid \calT_t]$ is upper bounded by $A_k\;\|x-c\|_2^2$. Observe, that $A_k$ is well defined and finite, because $\calT(x)$ and $\calT_t(x)$ take at most $k$ different values (namely, values in $\{c^1,\dots,c^k\}$).

We show an upper bound of $O(\nicefrac{1}{\varepsilon}\log k \log\log k)$ on $A_k$ (note: $\varepsilon = \min\{\nicefrac{\delta}{15\ln k}$,\nicefrac{1}{384}\}). To illustrate the proof, we make a number of simplifying assumptions in this section. The actual proof is considerably more involved. We give it in Section~\ref{sec:approx-factor}.

\medskip

\noindent\textbf{Informal Proof of the Upper Bound on $A_k$.}
Suppose $c^*$ is the tentative center for $x$ at step $t^*$. If at some step $t\geq t^*$, center $c^*$ is separated from $x$, then we assign a new tentative center to $x$. We call this center a \emph{fallback} center for $x$. This fallback center depends on the tree $\calT_t$ and cut $(i,\xi)$ that
separates $x$ and $c^*$. However, to illustrate the idea behind the proof, let us assume that the distance from the \emph{fallback} center to $x$ does not depend on the cut $(i,\xi)$. Specifically, we suppose that the distance from $x$ to the fallback center is $M_t$ at step $t$ \emph{for every cut} $(i,\xi)$. 

\medskip

\noindent We consider four possibilities:
\begin{enumerate}[A.]
\item Point $x$ and $c^*$ are never separated.
\item Point $x$ is separated from $c^*$ at step $t$ and $D_t^2 \leq \|x-c^*\|_2^2$.
\item Point $x$ is separated from $c^*$ at step $t$ and 
$\|x-c^*\|^2_2 < D_t^2 \leq A_k M_t^2/2$.
\item Point $x$ is separated from $c^*$ at step $t$ and 
$D_t^2 > A_k M_t^2/2$.
\end{enumerate}

In case (A), the cost of $x$ in the resulting tree $\calT$ equals $\|x-c^*\|_2^2$. In cases (B) and (C), the eventual cost of $x$ is upper bounded by $(D_t + \|x-c^*\|_2)^2\leq 2D_t^2 + 2\|x-c^*\|_2^2$ because no matter which center $c^{**}$ in $C_t$ is assigned to $x$ in $\calT$, the distance from $c^{**}$ to $x$ is at most 
$\|x-c^*\|_2 + \|c^*-c^{**}\|_2\leq \|x-c^*\|_2 +D_t$ (note: $D_t$ is the maximum distance between centers in $C_t$). Furthermore, in case~(B), $2D_t^2 + 2\|x-c^*\|^2 \leq 4\|x-c^*\|^2$. In case~(D), after step $t$, the distance from $x$ to the new tentative center is $M_t$. Hence, by the definition of $A_k$ (see Definition~\ref{def:Ak}), the 
expected cost of $x$ in $\calT$ is bounded by $A_k M_t^2$.
To summarize, in case (A) or (B), the final cost of $x$ is at most $4\|x-c^*\|_2^2$. In case (C) and (D), the final cost is upper bounded by 
$2\|x-c^*\|^2_2 + \min\Big\{2D^2_t, A_k M_t^2\Big\},$
where $t$ is the step when $x$ and $c^*$ are separated.

Let $t^{**}$ be the first step $t$ of the algorithm, when $D_t\leq \|x-c^*\|_2$ or $c^*$ is no longer the tentative center for $x$. Note that for some step $t$, $C_t$ contains only one center and $D_t = 0$. Hence, the stopping time $t^{**}$ is well defined.
Then,
\begin{multline*}
\E[\|x-\calT(x)\|_2^2\mid \calT_{t^*}]\leq
 4\|x-c^*\|^2_2+ \\ 
+ 
\E\bigg[
 \sum_{t= t^*}^{t^{**}-1}\pr\{x \;\&\;c^* \text{ are separated at step } t\mid \calT_{t}\} 
 \min\big\{2D^2_t, A_k M_t^2\big\}\mid \calT_{t^*}
\bigg].
\end{multline*}
We need to estimate the probability that $x$ and $c^*$ are separated at step $t$. 
Observe that if $x$ and $c^*$ are separated, then $x_i - m^t_i \leq  \sigma_t \sqrt{\theta_t}R_t$ and $c^*_i - m^t_i \geq (\sigma_t+\varepsilon) \sqrt{\theta}R_t$ or $x_i - m^t_i \geq \sigma_t \sqrt{\theta}R_t$ and $c^*_i - m^t_i \leq (\sigma_t-\varepsilon) \sqrt{\theta}R_t$, where $i=i_t$ is the coordinate chosen by the algorithm. We consider the case when $x_i$ and $c^*_i$ are on the same side of $m^t_i$, i.e. $(x_i - m^t_i)(c^*_i - m^t_i) \geq 0$. The case when $x_i$ and $c^*_i$ are on the opposite sides of $m^t_i$ is handled similarly. Since $\theta_t$ is uniformly distributed in $[0,1]$ and coordinate $i_t$ is chosen randomly from $\{1,\dots, d\}$, we have
\begin{multline*}
\pr\{x \;\&\;c^* \text{ are separated at step } t\mid \calT_{t}\} \leq \\ \leq \frac{1}{d\,R_t^2}\sum_{i=1}^{d}
\max\bigg\{\frac{|c^*_i - m^t_i|^2}{(1+\varepsilon)^2} - |x_i - m_i|^2,
|x_i - m_i|^2 - \frac{|c^*_i - m^t_i|^2}{(1-\varepsilon)^2},0\bigg\}.
\end{multline*}

\noindent\textbf{Remark:} In the formula above, we divide $|c^*_i - m^t_i|^2$ by $(1+\varepsilon)^2$ and $|c^*_i-m_i^t|^2$ by $(1-\varepsilon)^2$. These factors -- $\nicefrac{1}{(1+\varepsilon)^2}$ and 
$\nicefrac{1}{(1-\varepsilon)^2}$ -- are essential for the analysis. If we did not have them, we would get $\tilde\Theta(k)$ instead of 
$O(\nicefrac{1}{\varepsilon}\log k \log\log k)$ approximation!

\medskip

\noindent We now use the following inequality: For all positive numbers $a$, $b$ and $\varepsilon\in(0,1)$, we have
\begin{equation}\label{eq:quadratic-ab}
\max\bigg\{\frac{b^2}{(1+\varepsilon)^2} - a^2,
b^2 - \frac{a^2}{(1-\varepsilon)^2}\bigg\}\leq \frac{(b - a)^2}{2\varepsilon-\varepsilon^2}\leq \frac{(b - a)^2}{\varepsilon}.
\end{equation}
This inequality can be verified by dividing the left and right hand sides by $a^2$ and solving the obtained quadratic equation for $\lambda = b/a$. We have
$$
\pr\{x \;\&\;c^* \text{ are separated at step } t\mid \calT_{t}\} \leq \frac{1}{d\,R_t^2}\sum_{i=1}^{d}\frac{(x_i-c^*_i)^2}{\varepsilon}=\frac{\|x-c^*\|_2^2}{\varepsilon d\,R_t^2}.
$$
Note that the separation probability is proportional to the squared distance between $x$ and its tentative center $c^*$ (i.e., $\|x-c^*\|_2^2$) rather than the distance $\|x-c^*\|_2$ itself. 

In Section~\ref{sec:approx-factor}, we are going to use a slightly different version of  inequality~(\ref{eq:quadratic-ab}) to bound the  probability that $x$ and $c^*$ are separated using a particular cut $(i,\xi)$ (see Claim~\ref{cl:bound-eta}).

We use the upper bound on the separation probability to obtain a convenient bound on the expected final cost of $x$: 
\begin{align*}
\E[\|x-\calT(x)\|_2^2\mid \calT_{t^*}]&\leq
4\|x-c^*\|^2_2
+ 
\E\bigg[\sum_{t= t^*}^{t^{**}-1}\frac{\|x-c^*\|_2^2}{\varepsilon d R_t^2} \cdot 
 \min\big\{2D^2_t, A_k M_t^2\big\}\mid \calT_{t^*}
\bigg]\\
&= 
\|x-c^*\|^2_2\cdot\Bigg(4 + 
\E\bigg[\frac{1}{\varepsilon d}\sum_{t= t^*}^{t^{**}-1}
\frac{\min\big\{2D^2_t, A_k M_t^2\big\}}{R_t^2}
\mid \calT_{t^*}\bigg]\Bigg).
\end{align*}
Thus,
\begin{equation}\label{eq:informal-bound-A-k}
\E\bigg[\frac{\|x-\calT(x)\|_2^2}{\|x-c^*\|_2^2}\mid \calT_{t^*}\bigg]\leq
4 + 
\E\bigg[\frac{1}{\varepsilon d}\sum_{t= t^*}^{t^{**}-1}
\frac{\min\big\{2D^2_t, A_k M_t^2\big\}}{R_t^2}
\mid \calT_{t^*}
\bigg].
\end{equation}
Our goal is to bound the right hand side of this inequality by $O(\nicefrac{1}{\varepsilon}\log k \log\log k)$. 

In Lemma~\ref{lem:diamer-radius-bounds}, we show that $R_t\approx D_t$. Specifically, $\nicefrac{1}{\sqrt{2}} R_t \leq D_t \leq 2R_t$. This inequality would be trivial if $m^t$ was one of the centers $c^j$. However, generally speaking, this is not the case. In fact, $m^t$ does not have to belong to the convex hull of centers in $C_t$. Nevertheless, 
$D_t\in[\nicefrac{1}{\sqrt{2}} R_t, 2R_t]$ because $m^t$ is the median of $C_t$ (see Lemma~\ref{lem:diamer-radius-bounds}).

It is easy to see that the diameter $D_t$ is a non-increasing function of $t$ (since $C_{t+1}\subset C_t$) and $M_t$ is a non-decreasing function of $t$. In Lemma~\ref{lem:center-separate}, we show that, in fact, $D_t$ decreases by a factor of $2$ every $L = \Theta(d\ln k)$ steps with high probability. That is, $D_{t + L} \leq D_t/2$. This happens because for every step $t$, each pair of centers $c'$ and $c''$ with $\|c'-c''\|_2\geq D_t/2$ assigned to $u_t$ is separated with probability at least $\Omega(\nicefrac{1}{d})$ (see Corollary~\ref{cor:stop-time-t-end}). So, in $L=\Theta(d \ln k)$ steps all pairs of centers in $C_t$ at distance at least $D_t/2$ are separated with high probability.

We upper bound the right hand side of (\ref{eq:informal-bound-A-k}). Write
\begin{align}\label{eq:informal:min-two-terms}
\frac{1}{\varepsilon d}\sum_{t= t^*}^{t^{**}-1}
\frac{\min\big\{2D^2_t, A_k M_t^2\big\}}{R_t^2}&
\leq 
\sum_{\substack{t\in\{t^*,\cdots,t^{**}-1\}\\A_k M_t^2\leq 2D^2_t}}
\frac{A_k M_t^2}{\varepsilon d R_t^2}
+
\sum_{\substack{t\in\{t^*,\cdots,t^{**}-1\}\\2D^2_t < A_k M_t^2}}
\frac{2D^2_t}{\varepsilon d R_t^2}\notag
\\
&\leq 
\underbrace{
\sum_{\substack{t\in\{t^*,\cdots,t^{**}-1\}\\2D^2_t \geq A_k M_t^2}}
\frac{4A_k M_t^2}{\varepsilon d D_t^2}}_{\Sigma_{I}}+
\underbrace{
\sum_{\substack{t\in\{t^*,\cdots,t^{**}-1\}\\2D^2_t < A_k M_t^2}}
\frac{8}{\varepsilon d}}_{\Sigma_{II}}.
\end{align}
Consider the first sum, $\Sigma_{I}$ on the right hand side of (\ref{eq:informal:min-two-terms}). It is upper bounded by $2L$ times the maximum term in that sum, because $D_t$ halves every $L$ steps and therefore $(M_t/D_t)^2$ increases by 4 times every $L$ steps. The maximum term in $\Sigma_{I}$ is, in turn, upper bounded by $8/(\varepsilon d)$ (because $2D_t^2 \geq A_k M_t^2$ for all terms in $\Sigma_{I}$).

Now consider the second sum, $\Sigma_{II}$ on the right hand side of (\ref{eq:informal:min-two-terms}). Let $t'$ be the first step $t$ for which $2D_t^2 < A_k M_t^2$. Using that $D_{t + L} \leq D_t/2$, we obtain the following upper bound on the number of steps $t< t^{**}$ in $\Sigma_{II}$:
$$t^{**}-t'\leq 
L + L\cdot \log_2{\frac{D_{t'}}{D_{t^{**}-1}}}
\leq 
L + L\cdot \log_2{\frac{\sqrt{A_k/2}\; M_{t'}
}{D_{t^{**}-1}}}
\leq 
L + L\cdot \log_2{\frac{\sqrt{A_k/2}\; M_{t^{**}-1}
}{D_{t^{**}-1}}}.$$
The last inequality holds because $M_t$ is a non-decreasing function of $t$. Recall, that the distance to the fallback center, $M_t$ is upper bounded by $\|x-c^*\|_2 + D_{t}$
for every step $t\in \{t^*,\cdots, t^{**}-1\}$. Also,
by the definition of stopping time $t^{**}$, for every $t < t^{**}$, we have $D_t> \|x-c^*\|_2$. Thus,
$$\frac{M_{t^{**}-1}}{D_{t^{**}-1}}\leq 
\frac{\|x-c^*\|_2 + D_{t^{**}-1}}{D_{t^{**}-1}}\leq 2.$$
Therefore, $t^{**}-t' \leq L\cdot (1 + \log_2 \sqrt{2A_k})$. Consequently, the second sum, $\Sigma_{I}$ as well as $\Sigma_I+\Sigma_{II}$ are upper bounded by  $O((L\log A_k)/(\varepsilon d)) = O(\nicefrac{1}{\varepsilon} \log k \log A_k)$. 
We obtained the following bound:
$$\E\bigg[\frac{\|x-\calT(x)\|_2^2}{\|x-c^*\|_2^2}\mid \calT_{t^*}\bigg]\leq O(\nicefrac{1}{\varepsilon} \log k \log A_k).$$
Therefore, $A_k \leq O(\nicefrac{1}{\varepsilon} \log k \log A_k)$. This recurrence relation gives us an upper bound of $O(\nicefrac{1}{\varepsilon}\log k \log \log k)$ on $A_k$. This concludes the proof overview of Theorem~\ref{eq:thm-aprox-factor}.

\subsection{Expected Number of Leaves}
We show that the expected number of leaves in the threshold tree given by our algorithm is at most $e^{\delta/2}k$. Particularly, for $\delta \in (0,1)$, the expected number of leaves is at most $(1+\delta)k$. We now give an overview of the analysis. We provide a complete proof in  Section~\ref{sec:exp-number-leaves}.

In this section, we consider the case when the space is 1-dimensional. That is, all centers and data points lie on the real line. Consider a fixed center $c$. Let $N_c(\calT)$ be the number of leaves in tree $\calT$ containing $c$. We show that $\E[N_c(\calT)]$ is at most $e^{\delta/2}$. 

Suppose $c$ is assigned to node $u$ at step $t$ (note that $c$ may be assigned to several nodes). Denote the total number of centers assigned to $u$ by $k' = |C_u|$. We prove by induction on $k'$ that the expected number of leaves to which $u$ is assigned in the subtree rooted at $u$ is at most 
$(1+5\varepsilon)^{\log_2 k'}$. If $k'=1$, then the claim trivially holds, since $u$ is a leaf. Assume $k'>1$.

Our algorithm divides $u$ into two parts $u_{left}$ and $u_{right}$. One of them contains the median $m^u$. We call that part the main child and denote it by $u'$. In turn, the main child $u'$ is also divided into two parts, one of them -- denoted by $u''$ -- is the main child of $u'$. We call the sequence of nodes $u,u',u'',\dots$ the main branch rooted at $u$. Note that the main child always contains at least half of all centers assigned to its parent. This is the case, because $m^u$ is the median of all centers assigned to $u$. Thus, the part containing $m^u$ contains at least half of all centers in $C_u$, and the other (secondary) child contains at most half of all centers in $C_u$.

Suppose that center $c$ is assigned to a node $v$ in the main branch $u,u',u'',\dots$. When $v$ is divided into two parts, one of the following three events may occur: (1) $c$ is assigned only to the main child of $v$; (2) $c$ is assigned to both the main and secondary children of $v$; (3) $c$ is assigned only to the secondary child of $v$. Denote these events by $\calE_1$, $\calE_2$, and $\calE_3$, respectively. We estimate the number of nodes $w$ such that $c$ is assigned to $w$, and $w$ is a secondary child of a node in the main branch. This number equals to the number of events $\calE_2$ that occur in the main branch before the first event $\calE_3$ occurs plus 1. If the probabilities of events $\calE_1$, $\calE_2$, and $\calE_3$ were the same for all nodes in the main branch containing $c$, the expected number above would be equal to $1/\pr(\calE_3\mid \calE_2\cup \calE_3)$.
Without loss of generality assume that $m^u=0$, then for $\varepsilon \leq 1/10$, we have
$$
\frac{1}{\pr(\calE_3\mid \calE_2\cup \calE_3)} 
=
\frac{\pr(\calE_2\cup \calE_3)}{\pr(\calE_3)} 
=
\frac{c^2}{(1-\varepsilon)^2R_t^2}\left/
\frac{c^2}{(1+\varepsilon)^2R_t^2}\right. 
=
\frac{(1+\varepsilon)^2}{(1-\varepsilon)^2} \leq 1+5\varepsilon.
$$
Every secondary child $w$ contains at most $k'/2$ centers. So, by the inductive hypothesis, the expected number of leaves containing $c$ in the subtree rooted at $w$ is at most $(1+5\varepsilon)^{\floor{\log_2 k'/2}}$. Therefore, 
the expected number of leaves containing $c$ in the subree rooted at $u$ is at most 
$$(1+5\varepsilon)\cdot (1+5\varepsilon)^{\floor{\log_2 k'/2}} \leq (1+5\varepsilon)^{\floor{\log_2 k'}}.$$
This concludes the proof of the inductive claim. We now observe that 
$$\E[N_c(\calT)]\leq (1+5\varepsilon)^{\floor{\log_2 k}} \leq e^{\delta/2}$$
for $\varepsilon \leq \frac{\delta}{15\ln k}$.

%% file: paper/expected-leave-count.tex
\section{Expected Number of Leaves}\label{sec:exp-number-leaves}

In this section, we prove a bound the expected number of leaves in the threshold tree constructed by our algorithm. Our algorithm assigns all centers $c^1,\dots,c^k$ to the root $r$ of the threshold tree $\calT$. Then, it recursively divides 
centers assigned to every node $u$ between its children. However, centers in a narrow strip $Left \cap Right$ are shared by the both children of node $u$. Thus, the total number of leaves in the threshold tree $\calT$ may be larger than $k$. Let $N(\calT)$ be the number of leaves in $\calT$. We show an upper bound of $e^{\delta/2}k$ on the expected number of leaves $\E[N(\calT)]$, where the expectation is over the randomness of our algorithm. 

\begin{theorem}\label{thm:exp-num-leaves}
For every set of centers $c^1,c^2,\dots,c^k$ in $\bbR^d$ and every $\delta \in(0,\nicefrac{\ln k}{32})$, the expected number of leaves in the threshold tree $\calT$ given by our algorithm is at most
$$
\E_\calT[N(\calT)] \leq e^{\delta/2}k. 
$$
In particular, for $\delta \in (0,1)$, 
$$
\E_\calT[N(\calT)] \leq (1+\delta)k. 
$$
\end{theorem}

\begin{proof}
For every center $c$, we bound the expected number of leaves containing $c$ by $e^{\delta/2}$. Consider a fixed center $c$. For a node $u$ in the threshold tree $\calT$, let $N_c^u(\calT)$ denote the number of leaves in the subtree of $\calT$ rooted at node $u$ to which center $c$ is assigned to. 
\begin{definition}
For every integer $k' \in \{1,2,\dots,k\}$, let $B_{k'}$ be the minimum number such that the following inequality holds for every partially built tree $\calT_t$ and every leaf $u$ with $|C_u|\leq k'$ in $\calT_t$ to which center $c$ is assigned,
\begin{equation*}\label{eq:def:bk}
\E[N_c^u(\calT) \mid \calT_t]\leq B_{k'}.
\end{equation*}
\end{definition}
That is, $B_{k'}$ is an upper bound on the expected number of leaves in the subtree rooted at $u$ that contain $c$ if at most $k'$ centers are assigned to $u$. To prove Theorem~\ref{thm:exp-num-leaves}, it is sufficient to show that $B_k$ is at most $1+\delta$. We derive the following recurrence relation on $B_{k'}$.
\begin{lemma}\label{lem:leaves_recurrence}
The upper bound on the expected number of leaves $B_{k'}$ satisfies the following recurrence relation:
\begin{align}
    B_1 &= 1,\\
    B_{k'} &\leq (1+5\varepsilon) B_{\floor{k'/2}}, \label{eq:Bk-recurrence} 
\end{align}
where $\varepsilon = \min\{\nicefrac{\delta}{15\ln k} , 1/384\}$. 
\end{lemma}

\begin{proof}
It is easy to see that $B_1=1$, because if $c$ is the only center assigned to node $u$, then $u$ is a leaf and $N^u_c(\calT) = 1$.
We now prove~(\ref{eq:Bk-recurrence}). 
Consider a partially built tree $\calT_t$, node $u$ in $\calT_t$, and center $c$ in $X_u$ for which inequality (\ref{eq:def:bk}) is tight i.e., $B_{k'}=\E[N_c^u(\calT) \mid \calT_t]$. 

Examine the call of function \emph{Divide-and-Share} that splits node $u$. Let $i_t$ be the coordinate randomly chosen for this call of function \emph{Divide-and-Share}. Without loss of generality, we assume that $c_i \geq m^u_i$. 
If $\sigma_t$ is negative, then center $c$ is assigned only to the right child of $u$. In this case, the expected number of leaves containing $c$ in the subtree rooted at $u$ is at most $B_{k'}$.

We now consider the case when $\sigma_t = 1$. Define three disjoint events: (1) center $c$ is assigned only to the left child of $u$ and $\sigma_t = 1$; (2) center $c$ is assigned to both children of $u$ and $\sigma_t = 1$; (3) center $c$ is assigned only to the right child of $u$ and $\sigma_t = 1$. Denote these events by $\calE_1$, $\calE_2$, and $\calE_3$, respectively.

The number of centers assigned to node $u$ is $k'$. Thus, the number of centers assigned to each child of $u$ is at most $k'$. Moreover, if $\sigma_t = 1$, the number of centers assigned to the \emph{right} child $u_{right}$ of $u$ is at most $\floor{k'/2}$,  because $m^u$ is the median of all centers in $C_u$ and for all centers $c'$ assigned to $u_{right}$, $c'_i >  m^u_i$. Hence, if event $\calE_1$ occurs, then the expected number of leaves containing $c$ in the subtree rooted at $u$ is bounded by $B_{k'}$. If event $\calE_2$ occurs, then the expected number of leaves containing $c$ in the subtree rooted at $u$ is bounded by $B_{k'} + B_{\floor{k'/2}}$.  Finally, if event $\calE_3$ occurs, then the expected number of leaves containing $c$ in the subtree rooted at $u$ is bounded by $B_{\floor{k'/2}}$. Thus,
\begin{align*}
\E[N_c^u(\calT) \mid \calT_t] &\leq
\nicefrac{1}{2} B_{k'} + B_{k'}\pr(\calE_1\mid \calT_t)
+ \big( B_{k'}+ B_{\floor{k'/2}}\big)\pr(\calE_2\mid \calT_t) + B_{\floor{k'/2}} \pr(\calE_3\mid \calT_t) \\
&=
\Big((\nicefrac{1}{2} + \pr(\calE_1\mid \calT_t) + \pr(\calE_2\mid \calT_t)\Big) B_{k'} + \Big(\pr(\calE_2\mid \calT_t) + \pr(\calE_3\mid \calT_t)\Big) B_{\floor{k'/2}}.
\end{align*}
Since $\nicefrac{1}{2}+ \pr(\calE_1\mid \calT_t)+ \pr(\calE_2\mid \calT_t)+ \pr(\calE_3\mid \calT_t) = 1$, we have
$$B_{k'}= \E[N_c^u(\calT) \mid \calT_t] \leq
\Big((1 - \pr(\calE_3\mid \calT_t)\Big) B_{k'} + \Big(\pr(\calE_2\mid \calT_t) + \pr(\calE_3\mid \calT_t)\Big) B_{\floor{k'/2}}.
$$
Thus,
$$
B_{k'}\leq \frac{\pr(\calE_2\cup \calE_3\mid \calT_t)}{\pr(\calE_3\mid \calT_t)}B_{\floor{k'/2}}.
$$
Compute $\pr(\calE_2\cup \calE_3\mid \calT_t)$ and $\pr(\calE_3\mid \calT_t)$:
\begin{align*}
\pr(\calE_2\cup\calE_3\mid \calT_t) & = \frac{1}{2d} \sum_{i=1}^d \prob{\abs{c_i -m_i^t} \geq (1-\varepsilon)\sqrt{\theta_t}R_t}= \frac{1}{2d} \sum_{i=1}^d
\frac{(c_i - m_i^t)^2}{(1-\varepsilon)^2R_t^2};\\
\pr(\calE_3\mid \calT_t) & = \frac{1}{2d} \sum_{i=1}^d \prob{\abs{c_i -m_i^t} \geq (1+\varepsilon)\sqrt{\theta_t}R_t} =\frac{1}{2d} \sum_{i=1}^d\frac{(c_i - m_i^t)^2}{(1+\varepsilon)^2R_t^2}.
\end{align*}
Therefore, we have
$$
B_{k'} \leq 
B_{\floor{k'/2}}\cdot \sum_{i=1}^d\frac{(c_i - m_i^t)^2}{(1-\varepsilon)^2R_t^2}\left/\sum_{i=1}^d\frac{(c_i - m_i^t)^2}{(1+\varepsilon)^2R_t^2}\right.
= \frac{(1+\varepsilon)^2}{(1-\varepsilon)^2} B_{\floor{k'/2}} \leq (1+5\varepsilon) B_{\floor{k'/2}}.
$$
where the last inequality holds because $\varepsilon \leq 1/10$.
\end{proof}

We now bound the expected number of leaves in the threshold tree $\calT$. By Lemma~\ref{lem:leaves_recurrence}, the expected number of leaves containing center $c$ in the threshold tree $\calT$ is at most
$$
\E[N_c^r(\calT)] \leq B_k \leq (1+5\varepsilon)^{\floor{\log_2 k}} \cdot B_1 \leq
\Big(1+\frac{\delta}{3\ln k}\Big)^{\log_2 k} 
\leq  \big(e^{\frac{\delta}{3\ln k}}\big)^{\log_2 k}
< e^{\delta/2}.
$$
Since $e^{\delta/2} < 1 + \delta$ for $\delta \in(0, 1)$, we have for $\delta \in (0,1)$
$$
\E[N_c^r(\calT)] \leq e^{\delta/2} \leq 1 + \delta.
$$
\end{proof}

%% file: paper/approx-factor.tex
\section{Approximation Factor}\label{sec:approx-factor}
We now prove Theorem~\ref{thm:main-approx-factor}. Our proof follows the outline given in Section~\ref{sec:proof-overview}. We fix a point $x$, step $t^*$, and estimate $\E[\|x - \calT(x)\|_2^2 \mid \calT_{t^*}]$. Let $c^*=\calT_{t^*}(x)$ be the tentative center assigned to $x$ at step $t^*$. As in Section~\ref{sec:proof-overview}, let $u_t$ be the leaf node of $\calT_t$ that contains $x$, $C_t=C_{u_t}$, $R_t = R_{u_t}$, and $m^t=m^{u_t}$. We denote the diameter of $C_t$ by $D_t$. 

\subsection{Bounds on the Diameter}
We prove several facts about the diameter $D_t$. First, we show that $D_t\approx R_t$.

\begin{lemma}\label{lem:diamer-radius-bounds}
For every leaf node $u$ in a partially built tree $\calT_t$, we have $$\nicefrac{1}{\sqrt{2}}R_u \leq D_{u}\leq 2R_{u}.$$
\end{lemma}
\begin{proof}
The second bound easily follows from the triangle inequality: for every $c'$ and $c''$ in $C_u$,
$$\|c'-c''\|_2 \leq \|c'-m^u\|_2+\|m^u-c''\|_2 \leq 2R_u.$$
We now show the first bound. Let $c$ be the farthest center in $C_u$ from $m^{u}$. Then, $R_u = \|c-m^u\|_2$. Consider a center $c'$ in $C_u$. The distance between $c$ and $c'$ is upper bounded by $D_u$ because $D_u$ is the diameter of $C_u$. Hence, for each $c'$ in $C_u$, we have 
$\|c - c'\|_2^2\leq D_u^2$. Thus,
$$D_u^2 \geq \Avg_{c'\in C_u} \|c - c'\|_2^2 = \Avg_{c'\in C_u} \sum_{i=1}^{d} |c_i - c'_i|^2
= \sum_{i=1}^{d} \Avg_{c'\in C_u} |c_i - c'_i|^2,$$
where $\Avg_{c'\in C_u} f(c')$ denotes the average of $f$ over $c'$ in $C_u$. Observe that 
$$\Avg_{c'\in C_u} |c_i - c'_i|^2\geq \nicefrac{1}{2}|c_i-m^u_i|^2.$$
This is because $m^u$ is the median point in $C_u$, consequently, at least a half of all points $c'\in C_u$ are on the other side of the hyperplane $\{x: x_i = m^u_i\}$ from $c$ (including centers $c'$ on the hyperplane). For these centers $c'$, we have $|c_i-c'_i| \geq |c_i-m^u_i|$. Therefore,
$$D_u^2 \geq \sum_{i=1}^{d} \Avg_{c'\in C_u} |c_i - c'_i|^2 \geq \nicefrac{1}{2}\sum_{i=1}^{d}|c_i-m^u_i|^2 = 
\nicefrac{1}{2} R_u^2.
$$
\end{proof}

We prove that the diameter $D_t$ is exponentially decaying with $t$. To this end, we estimate the probability that two centers $c'$ and $c''$ with $\|c'-c''\|_2 \geq D_t/2$ are separated at step $t$. We say that two centers $c',c'' \in C_t$ are separated at step $t$ if $c' \notin C_t$ or $c'' \notin C_t$. 


\begin{lemma}\label{lem:center-separate}
For every two centers $c',c'' \in C_t$ at distance at least $D_t/2$, 
$$\Pr\Big\{c'\notin C_{t+1} \text{ or } c''\notin C_{t+1}\mid \calT_t\Big\}\geq \nicefrac{1}{128d}.$$
\end{lemma}
\begin{proof}
Suppose, at step $t$, the algorithm picks coordinate $i_t=i$. For every two centers $c',c'' \in C_t$, we consider the following two cases: (1) $c'$ and $c''$ are on the same side of the median $m^t$ in coordinate $i$ (i.e. $\sign(c'_i-m^t_i) = \sign(c''_i-m^t_i)$), and (2) $c'$ and $c''$ are on the opposite sides of the median $m^t$ in coordinate $i$ (i.e. $\sign(c'_i-m^t_i) = -\sign(c''_i-m^t_i)$).

Consider the first case, when $c'$ and $c''$ are on the same side of the median $m^t$ in coordinate $i$. Without loss of generality, assume that $c''_i\geq c'_i \geq m^t_i$. Observe that if
$\sigma_t = 1$, $c''_i -m^t_i > (1+\varepsilon)R_t \sqrt{\theta_t}$, and  
$c'_i -m^t_i \leq (1-\varepsilon)R_t \sqrt{\theta_t}$, then centers $c'$ and $c''$ are separated at step $t$. 
Let $\calE_{t,i,c'} = \{i_t=i, \sigma_t = 1\}$ be the event that the threshold cut at step $t$ is in coordinate $i$ and $\sigma_t =1$. Then, the conditional probability that $c'$ and $c''$ are separated  given $\calE_{t,i,c'}$ is \begin{align*}
\pr\Big[c'_i -m^t_i\leq (1-\varepsilon)R_t\sqrt{\theta_t}&
\text{ \& } 
c''_i -m^t_i> (1+\varepsilon) R_t\sqrt{\theta_t}
\mid \calT_t,\;\calE_{t,i,c'}\Big]\\
&=\pr\Big\{\theta_t \in \Big[\frac{(c'_i-m^t_i)^2}{(1-\varepsilon)^2R_t^2}, 
\frac{(c''_i-m^t_i)^2}{(1+\varepsilon)^2R_t^2}\Big]\Big\} \\
 &= \Bigg(
\frac{(c''_i-m^t_i)^2}{(1+\varepsilon)^2R_t^2}  -\frac{(c'_i-m^t_i)^2}{(1-\varepsilon)^2R_t^2}\Bigg)^+,
\end{align*}
where $(x)^+$ denotes $\max\{x,0\}$.

Now, consider the second case, when $c'$ and $c''$ are on the opposite sides of the median $m^u$ in coordinate $i$. Assume without loss of generality that $c''_i \geq m^t_i \geq c'_i$ and $|c''_i -m^t_i| \geq |c'_i -m^t_i|$. If $c''_i-m^u_i \geq (1+\varepsilon)R_t\sqrt{\theta_t}$ and $\sigma_t = 1$, then $c'$ and $c''$ are separated at this step.
Thus, the conditional probability that $c'$ and $c''$ are separated given $i_t= i$ and parameter $\sigma_t = 1$ is at least
$$
\prob{c''_i-m^t_i \geq (1+\varepsilon)R_t\sqrt{\theta_t}\mid \calT_t,i_t=i, \sigma_t = 1} = \frac{(c''_i-m^t_i)^2}{(1+\varepsilon)^2R_t^2}.
$$

Define 
$$
a_i = \min\big\{|c'_i-m^t_i|,|c''_i-m^t_i|\big\}
\;\;\text{ and }\;\;
b_i = \max\big\{|c'_i-m^t_i|,|c''_i-m^t_i|\big\}.
$$
Let $I_1,I_2 \subset \{1,2,\dots, d\}$ be the set of indices $i$ for which $c'_i$ and $c''_i$ lie on the same side and opposite sides of $m^t$, respectively. Then, 
\begin{multline*}
\Pr\Big\{c'\notin C_{t+1} \text{ or } c''\notin C_{t+1}\mid \calT_t\Big\}
\geq\\\geq
\frac{1}{2d} \sum_{i \in I_1} \Big(
\frac{b_i^2}{(1+\varepsilon)^2R_t^2}  -\frac{a_i^2}{(1-\varepsilon)^2R_t^2}\Big)^+ +  \frac{1}{2d}\sum_{i \in I_2}\frac{b_i^2}{(1+\varepsilon)^2R_t^2}.
\end{multline*}

Now observe that
\begin{align*}
\frac{1}{2d}\sum_{i \in I_1}\Big(
\frac{b_i^2}{(1+\varepsilon)^2R_t^2}  -&\frac{a_i^2}{(1-\varepsilon)^2R_t^2}\Big)^+ \geq
\\&\geq
\frac{1}{2d R_t^2}\sum_{i \in I_1}
\frac{b_i^2}{(1+\varepsilon)^2}  -\frac{a_i^2}{(1-\varepsilon)^2} \\
&\geq 
\frac{1}{2d R_t^2}
\sum_{i \in I_1} b_i^2-a_i^2 -  (2\varepsilon b_i^2 + 3\varepsilon a_i^2).
\end{align*}
Similarly, we have
$$
\frac{1}{2d}\sum_{i \in I_2}\frac{b_i^2}{(1+\varepsilon)^2 R_t^2} \geq \frac{\sum_{i \in I_2} b_i^2 - 2\varepsilon b_i^2}{2R_t^2 d}.
$$
When $c'$ and $c''$ are on the same side of $m^t$ in coordinate $i$, we have 
$$
b_i^2 - a_i^2 = (b_i-a_i)(b_i + a_i) \geq (b_i-a_i)^2 = (c'_i-c''_i)^2. 
$$
When $c'$ and $c''$ are on the opposite side of $m^t$ in coordinate $i$, we have 
$$
4b_i^2 \geq (b_i+a_i)^2 =
(c'_i-c''_i)^2.
$$
Note that $\sum_{i=1}^d b_i^2 + a_i^2 = \sum_{i=1}^d (c'_i-m^t_i)^2 + (c''_i-m^t_i)^2 = \norm{c'-m^t}_2^2 + \norm{c''-m^t}_2^2$. Therefore, the probability that $c'$ and $c''$ are separated at step $t$ is at least
\begin{align*}
\prob{c'\notin C_{t+1} \text{ or } c''\notin C_{t+1} \mid \calT_t} &\geq \sum_{i=1}^{d} \frac{(c'_i-c''_i)^2}{8dR_t^2} - \frac{2\varepsilon b_i^2 + 3\varepsilon a_i^2}{2dR_t^2} \\
&\geq \frac{\norm{c'-c''}_2^2}{8R_t^2 d} - \frac{6\varepsilon}{2d}.
\end{align*}
where the second inequality is due to $\sum_{i=1}^d 2\varepsilon b_i^2 + 3\varepsilon a_i^2 \leq \sum_{i=1}^d 3\varepsilon b_i^2 + 3\varepsilon a_i^2 \leq 3\varepsilon\norm{c'-m^t}_2^2 + 3\varepsilon\norm{c''-m^t}_2^2 \leq 6\varepsilon R_t^2$.
We conclude that for centers $c'$ and $c''$ with $\|c'-c''\|_2^2\geq D_t^2/4$, we have
\begin{align*}
\prob{c'\notin C_{t+1} \text{ or } c''\notin C_{t+1} \mid \calT_t}&\geq \frac{1}{2d}\cdot \Big(\frac{D_t^2}{16R_t^2}-5\varepsilon\Big)\\ &\geq 
\frac{1}{2d}\cdot \Big(\frac{1}{32}-5\varepsilon\Big)\geq
\frac{1}{128 d}.
\end{align*}
Here, we used that $D_t \geq \nicefrac{1}{\sqrt{2}}R_t$ and $\varepsilon \leq \nicefrac{1}{384}$.
\end{proof}

We obtain the following corollary from Lemma~\ref{lem:diamer-radius-bounds}.

\begin{lemma}\label{lem:prob-long-phase}
Let $L=\ceil{640 d \ln k}$. Then, for every $t$, we have 
$$\prob{D_{t+ L} \geq D_t/2\mid \calT_t}\leq \frac{1}{k^3}.$$
\end{lemma}
\begin{proof}
Consider a fixed time step $t$. Suppose the distance  between centers $c'$ and $c''$ is at least $D_t/2$. Since the diameter $D_t$ is non-increasing as $t$ increases, the distance between $c'$ and $c''$ is greater than $D_{t'}/2$ for any step $t' \geq t$. By Lemma~\ref{lem:center-separate}, the probability that these centers $c'$ and $c''$ are separated at step $t'$ is at least $\nicefrac{1}{128d}$.

Thus, these two centers $c'$ and $c''$ are not separated in $\ceil{640d\ln k}$ steps from step $t$ with probability at most
$$
\left(1-\frac{1}{128d}\right)^{640d \ln k} \leq e^{-5\ln k}.
$$
Since there are at most $\binom{k}{2}$ pairs of centers with distance greater than $D_t/2$, by the union bound over all such pairs, we have for $L = \ceil{640 d \ln k}$
$$ 
\prob{D_{t+L} \geq D_t/2 \mid \calT_t} \leq \binom{k}{2}\cdot e^{-5\ln k} \leq \frac{1}{k^3}. 
$$
\end{proof}

To simplify the exposition, we define a stopping time $t^{**}$. Let $t^{**}$ be the first step $t > t^*$ of the algorithm when one of the following happens: (A) $D_t\leq \|x-c^*\|_2$ (note: if $c^*$ is the only center remaining in $C_t$, then $D_t = 0$); (B) $x$ and $c^*$ are separated before step $t$ (i.e., $c^*\notin C_t$); or
(C) $D_t > D_{t-L'}/2$ and $t \geq t^* + L'$ for $L' = \ceil{1280 d \ln k}$. For some step $t$, $C_t$ contains only one center and $D_t = 0$. Thus, the stopping time $t^{**}$ is well-defined. We show that it is very unlikely that the case (C) happens, i.e. $D_{t^{**}} > D_{t^{**}-L'}/2$ and $t^{**} \geq t^* + L'$.

\begin{corollary}\label{cor:stop-time-t-end}
Let $L'=\ceil{1280 d \ln k}$ be twice as large as $L$ in  Lemma~\ref{lem:prob-long-phase}. Then, 
$$\prob{D_{t^{**}} > D_{t^{**}-L'}/2\;\&\;t^{**}\geq t^*+L' \mid \calT_{t^*}}\leq \frac{1}{k}.$$
\end{corollary}
\begin{proof}
Let $L = \ceil{640 d \ln k}$ be as in Lemma~\ref{lem:prob-long-phase}. We consider the set of steps 
$$
S_L = \{t \leq t^{**} : t = t^* + Lz, z \geq 1 \}.
$$
By Lemma~\ref{lem:prob-long-phase}, we have for each step $t = t^* + Lz$ in this set $S_L$ 
$$
\prob{D_{t} > D_{t-L}/2 \mid \calT_{t-L}} \leq \frac{1}{k^3}.
$$
We consider every step $t = t^* + L'z$ for $z \geq 1$. If $D_t > D_{t-L'}/2$, then we have $t^{**} \leq t$. If $D_t \leq D_{t-L'}/2$, then we must separate at least one center from $C_{t-L'}$ in $L'$ steps, which means 
$|C_{t}| < |C_{t-L'}|$. Since there are at most $k$ centers in $C_{t^*}$, we have at most $k$ such steps $t$ with $D_t \leq D_{t-L'}/2$. Thus, we have $t^{**} \leq t^*+L' k = t^* + 2k L$. Then, the set of steps $S_L$ contains at most $2k$ steps. By the union bound over all steps $t \in S_L$, we have $D_{t} \leq D_{t-L}/2$ for all steps $t \in  S_L$ with probability at least $1-1/k$. 
Suppose that $D_{t} \leq D_{t-L}/2$ holds for all steps $t \in  S_L$. For every $t^* + L' \leq t \leq t^{**}$, there exists a $t' \in S_L$ such that $t-L' \leq t'-L < t' \leq t$. Since $D_t$ is a non-increasing sequence, we have for every $t^* + L' \leq t \leq t^{**}$
$$
D_t \leq D_{t'} \leq D_{t'-L}/2 \leq D_{t-L'}/2.
$$
Therefore, we have $D_{t^{**}} > D_{t^{**}-L'}/2$ and $t^{**} \geq t^* + L'$ with probability at most $1/k$. 
\end{proof}

\subsection{Cost of Separation}
In this section, we complete the proof of Theorem~\ref{thm:main-approx-factor}. The proof is similar to the overview we gave in Section~\ref{sec:proof-overview}. The key difference is that we no longer assume that the distance from $x$ to the nearest fallback center does not depend on the cut that separates $x$ and $c^*$. 

To simplify the exposition, from now on, we shall assume that $c^*_i\geq x_i$ for all $i$. We make this assumption without loss of generality, because if $c^*_i < x_i$ for some $i$, we can
mirror all centers $c$ in $C$ and point $x$ across the hyperplane $\{y_i = 0\}$, or, in other words, we can change the sign of the $i$-th coordinate for all centers $c$ in $C$ and point $x$. This transformation does not affect the algorithm but makes $c^*_i \geq x_i$.

For every $(i,\eta)$ with $x_i\leq \eta < c_i$, define  $M_t(i,\eta)$ as follows: $M_t(i,\eta)$ equals the distance from $x$ to the closest center $c'$ in $C_t$ with $c'_i\leq \eta$. 
If there are no centers $c'$ in $C_t$ with $c'_i\leq \eta$, then we let $M_t(i,\eta)= \infty$. 
Observe that if $x$ and $c^*$ are separated at step $t$, then 
$$x_i\leq m^t_i + \sigma_t \sqrt{\theta_t}R_t
< 
\underbrace{m^t_i + \sigma_t \sqrt{\theta_t}R_t + \varepsilon \sqrt{\theta_t}R_t}_{\eta_t} < c^*_i,$$
where $i$ is the coordinate chosen at step $t$. Thus, if  $x$ and $c^*$ are separated at step $t$, the distance from 
$x$ to the fallback center is $M_t(i,\eta_t)$, where $\eta_t = m^t_i + \sigma_t \sqrt{\theta_t}R_t + \varepsilon \sqrt{\theta_t}R_t$.

At each step $t$, our algorithm calls function \emph{Divide-and-Share} with parameters $(i_t,\sigma_t,\theta_t)$ to split node $u_t$.  Let $\omega_t = (i_t,\xi_t)$ be the cut chosen by the algorithm for node $u_t$ where $\xi_t = m^t_i + \sigma_t \sqrt{\theta_t}R_t$; $\omega_t$ is undefined ($\omega_t=\perp$), if the algorithm does not make any cut at step $t$. Note that the cut $\omega_t$ is determined by the tuple $(i_t,\sigma_t,\theta_t)$. Then, $x$ and $c^*$ are separated at step $t$ by the tuple $(i,\sigma,\theta)$ if  $c^* \in C_t$, $\omega_t = (i, m^t_i + \sigma \sqrt{\theta}R_t)$ and $x_i \leq \xi_t < \eta_t < c^*_i$.

We define a penalty function $Z_t(i,\sigma,\theta)$ for every tuple $(i,\sigma,\theta)$ with $i \in \{1,2\dots,d\}, \sigma \in \{\pm 1\},\theta \in (0,1)$ as follows:
$$Z_t(i,\sigma,\theta) =
\begin{cases}
\E\big[\|x-\calT(x)\|_2^2\mid \calT_t,\; \omega_t = (i, m^t_i + \sigma \sqrt{\theta}R_t)\big],&\text{if $(i,\sigma,\theta)$ separates $x$ \& $c^*$ at step $t$};\\
0,&\text{otherwise}.
\end{cases}
$$
In other words, $Z_t(i,\sigma,\theta)$ equals $0$ if the tuple $(i,\sigma,\theta)$ does not separate $x$ and $c^*$ at step $t$. Otherwise, it is equal to the expected cost of $x$ in the final tree $\calT$ assuming that the algorithm chooses the tuple $(i,\sigma,\theta)$ at step $t$. Note that if $x$ and $c^*$ are already separated at step $t$, then $Z_t(i,\sigma,\theta) = 0$.

\begin{claim}\label{cl:two-bounds-on-Z}
For every step $t$ and every tuple $(i,\sigma,\theta)$, we have
$$Z_{t}(i,\sigma,\theta) \leq \min\big\{2\|x-c^*\|_2^2+2D^2_t,A_k\; M_t^2(i, \eta)\big\},$$
where $\eta = m^t_i + (\sigma+\varepsilon) \sqrt{\theta}R_t$.
\end{claim}
\begin{proof}
If $x$ and $c^*$ are not separated by the tuple $(i,\sigma,\theta)$ at step $t$ or $x$ and $c^*$ are already separated at step $t$, then we have $Z_t(i,\sigma,\theta) = 0$. Thus, we only need to consider the case when $x$ and $c^*$ are separated by the tuple $(i,\sigma,\theta)$ at step $t$. By the triangle inequality, we have
$$
\|x-\calT(x)\|_2^2 \leq (\|x-c^*\|_2 + \|c^*-\calT(x)\|_2)^2 \leq (\|x-c^*\|_2 + D_t)^2 \leq 2\|x-c^*\|_2 + 2D_t^2. 
$$
By Definition~\ref{def:Ak} of the approximation factor $A_k$, we have
$$
Z_t(i,\sigma,\theta) = \E\Big[\|x-\calT(x)\|_2^2\mid \calT_t,\; \omega_t = (i,m^t_i + \sigma \sqrt{\theta}R_t)\Big] \leq A_k 
\norm{x- \calT_{t+1}(x)}_2^2 = A_k M_t^2(i,\eta).
$$
Combining these two bounds, we get the conclusion.
\end{proof}

Our goal is to show that $A_k\leq O(\nicefrac{1}{\varepsilon}\log k \log \log k)$. We prove Lemma~\ref{lem:expected-cost-in-Ak}, which
provides the following recurrence relation on $A_k$: 
$A_k \leq \max\{4, \nicefrac{A_k}{k}\} + \nicefrac{\alpha}{\varepsilon} \log k \log A_k$.
Using this recurrence relation, we get the desired bound on $A_k$. 

\begin{lemma}\label{lem:expected-cost-in-Ak}
For some absolute constant $\alpha$, we have
\begin{equation}\label{eq:lem:expected-cost-in-Ak}
\E\bigg[\frac{\|x-\calT(x)\|_2^2}{\|x-c^*\|_2^2} \mid \calT_{t^*}\bigg]\leq \max\{4, \nicefrac{A_k}{k}\} + \nicefrac{\alpha}{\varepsilon} \log k \log A_k.    
\end{equation}
\end{lemma}
\begin{proof}
Let $t^{**}$ be the stopping time from Corollary~\ref{cor:stop-time-t-end}: $t^{**}$ is the first step $t$ when (A) $D_t\leq \|x-c^*\|_2$ (note: if $c^*$ is the only center remaining in $C_t$, then $D_t = 0$); (B) $x$ and $c^*$ are separated before step $t$ (i.e., $c^*\notin C_t$); or
(C) $D_t > D_{t-L'}/2$ (where $L'=O(d\ln k)$ as in Corollary~\ref{cor:stop-time-t-end}; $t\geq t^*+L'$). Let $\calE_A$, $\calE_B$, and $\calE_C$ be events corresponding to the the stopping rules (A), (B), and (C):
\begin{align*}
\calE_A &= \big\{D_{t^{**}} \leq \|x-c^*\|_2 \;\&\; c^*\in C_{t^{**}}\big\};\\
\calE_B &= \big\{x\;\&\;c^* \text{ are separated at step } t^{**} - 1\big\};\\
 \calE_C &= \big\{D_{t^{**}} > D_{t^{**}-L'}/2\;\&\;t^{**}\geq t^*+L'\big\}\setminus (\calE_A\cup \calE_B).
\end{align*}
Note that $\calE_A$, $\calE_B$, and $\calE_C$ are disjoint collectively exhaustive events (one of them must always occur) and by Corollary~\ref{cor:stop-time-t-end},
$\Pr(\calE_C\mid \calT_{t^*})\leq 1/k$. We further partition $\calE_B$ into disjoint events 
$$\calE_{B,t} = \{x\;\&\;c^* \text{ are separated at step } t\}.$$

If event $\calE_A$ occurs, then the eventual cost of $x$ is at most $(\|x-c^*\|_2 + D_{t^{**}})^2\leq 4 \|x-c^*\|_2^2$ because every center in $C_{t^{**}}$ is at distance at most 
$\|x-c^*\|_2 + D_{t^{**}}$ from $x$. If event $\calE_{B,t}$ occurs, then the expected cost of $x$ is upper bounded by $Z(i_t,\sigma_t,\theta_t)$. 
Finally, if event $\calE_C$ occurs, then the expected cost of $x$ in $\calT$ is upper bounded by $A_k \|x - c^*\|_2^2$ (because $c^*$ is the tentative center for $x$ at step $t^{**}$). We have
\begin{align*}
\E\big[\|x-\calT(x)\|_2^2\mid \calT_{t^*}\big] &
\leq 4\|x-c^*\|_2^2\cdot  \pr(\calE_A\mid \calT_{t^*})+A_k\|x-c^*\|_2^2\cdot \pr(\calE_C\mid \calT_{t^*})\\
&\phantom{\leq}+ \sum_{t= t^*}^{\infty} \E\Big[Z_t(i_t,\sigma_t,\theta_t) \mid \calE_{B,t},\; \calT_{t^*}\Big]
\Pr(\calE_{B,t}\mid \calT_{t^*})\\
&\leq \max\{4, A_k/k\} \cdot \|x-c^*\|_2^2
+ \sum_{t=t^*}^{\infty} \E\Big[(Z_t(i_t,\sigma_t,\theta_t) - 4\|x-c^*\|_2^2) \cdot \one (\calE_{B,t})\mid \calT_{t^*}\Big]
.
\end{align*}
Let $\widetilde{Z}_t(i_t,\sigma_t,\theta_t) = \max\{Z_t(i_t,\sigma_t,\theta_t) - 4\|x-c^*\|_2^2,0\}$. Then, 
$$\E\bigg[\frac{\|x-\calT(x)\|_2^2}{\|x-c^*\|_2^2} \mid \calT_{t^*}\bigg]\leq \max\{4, \nicefrac{A_k}{k}\} + \sum_{t=t^*}^{\infty} \E\bigg[\frac{\widetilde Z_t(i_t,\sigma_t,\theta_t)}{ \|x-c^*\|_2^2} \cdot \one (\calE_{B,t})\mid \calT_{t^*}\bigg]
. $$
Our goal is to upper bound the second term by 
$\nicefrac{\alpha}{\varepsilon}\log k\log A_k$.
Write,
\begin{equation}\label{eq:Z-bound}
\E\bigg[\widetilde Z_t(i_t,\sigma_t,\theta_t) \cdot \one (\calE_{B,t})\mid \calT_{t^*}\bigg] = 
\sum_{i=1}^{d}
\E\Bigg[
\int_0^1 \frac{\widetilde Z_t(i,-1,\theta)
+ \widetilde Z_t(i,1,\theta)}{2d}\;d\theta\cdot \one\{t < t^{**}\}\mid \calT_{t^*}\Bigg].
\end{equation}
Here, we used that parameters $i_t$, $\sigma_t$, and $\theta_t$ are randomly chosen from $\{1,\dots,d\}$, $\{\pm 1\}$, and $[0,1]$, respectively. We need the following lemma, which we prove in Section~\ref{sec:lem:subst-vars}.
\begin{lemma}\label{lem:subst-vars} For every $i$, we have
$$\int_0^1 \frac{\widetilde Z_t(i,-1,\theta)
+ \widetilde Z_t(i,1,\theta)}{2}\;d\theta
\leq \frac{c_i^*-x_i}{\varepsilon(1-\varepsilon)}\; \int_{x_i}^{c^*_i} \frac{\min\{2D_t^2,A_kM_t^2(i,\eta)\}}{ R_t^2} \;d\eta.$$
\end{lemma}
Using Lemma~\ref{lem:subst-vars}, we can upper bound~(\ref{eq:Z-bound}) as follows
\begin{align*}
\E\bigg[\widetilde Z_t(i_t,\sigma_t,\theta_t) \cdot \one (\calE_{B,t})\mid \calT_{t^*}\bigg] 
&\leq
\frac{1}{d} \sum_{i=1}^{d} \frac{c_i^*-x_i}{\varepsilon(1-\varepsilon)}\;
\E\bigg[
\sum_{t=t^*}^{t^{**}-1}\int_{x_i}^{c^*_i}
\frac{\min\{2D_t^2,A_kM_t^2(i,\eta)\}}{ R_t^2} \;d\eta
\mid \calT_{t^*}\bigg]\\
&=
\frac{1}{d} \sum_{i=1}^{d} \frac{c_i^*-x_i}{\varepsilon(1-\varepsilon)}\;
\int_{x_i}^{c^*_i}
\E\bigg[
\sum_{t=t^*}^{t^{**}-1}\frac{\min\{2D_t^2,A_kM_t^2(i,\eta)\}}{ R_t^2}
\mid \calT_{t^*}\bigg]\;d\eta\\
&\leq
\frac{1}{d} \sum_{i=1}^{d} \frac{2(c_i^*-x_i)^2}{\varepsilon}\;
\max_{\eta\in[x_i,c_i^{*}]}
\E\bigg[
\sum_{t=t^*}^{t^{**}-1}
\frac{\min\{2D_t^2,A_k M_t^2(i,\eta)\}}{ R_t^2}
\mid \calT_{t^*}\bigg]
.
\end{align*}
We now show that for every $\eta\in[x_i,c^*_i]$ the following bound holds with probability 1:
\begin{equation}\label{eq:sum-less-d-logk-loglogk}
\sum_{t=t^*}^{t^{**}-1}
\frac{\min\{2D_t^2,A_k M_t^2(i,\eta)\}}{ R_t^2} 
\leq O(d\log k \log A_k).
\end{equation}
This will conclude the proof of Lemma~\ref{lem:expected-cost-in-Ak} because (\ref{eq:sum-less-d-logk-loglogk}) implies that
$$
\E\bigg[\widetilde Z_t(i_t,\sigma_t,\theta_t) \cdot \one (\calE_{B,t})\mid \calT_{t^*}\bigg] \leq 
\frac{1}{d} \sum_{i=1}^{d} \frac{2(c_i^*-x_i)^2}{\varepsilon}\cdot O(d \log k \log A_k) = \frac{2\|c^*-x\|_2^2}{\varepsilon}\cdot O(\log k \log A_k).
$$
\end{proof}
\begin{lemma}\label{lem:eq:sum-less-d-logk-loglogk}
Inequality~(\ref{eq:sum-less-d-logk-loglogk}) holds with probability $1$.
\end{lemma}
\begin{proof}
By Lemma~\ref{lem:diamer-radius-bounds}, $R_t \geq D_t/2$. Thus,
$$\sum_{t=t^*}^{t^{**}-1}
\frac{\min\{2D_t^2,A_k M_t^2(i,\eta)\}}{ R_t^2}\leq 
8\sum_{t=t^*}^{t^{**}-1}
\frac{\min\{D_t^2,A_k M_t^2(i,\eta)\}}{D_t^2}=
8\sum_{t=t^*}^{t^{**}-1}
\min\Big\{1,\frac{A_k\,M_t^2(i,\eta)}{{D_t^2}}\Big\}.
$$
Let
$$f_t(i,\eta) = \frac{A_k\,M^2_t(i,\eta)}{D_t^2}.$$
Observe that $M_t(i,\eta)$ is a non-decreasing sequence and $D_t$ is a non-increasing sequence for fixed $i$, $\eta$ and $t\in\{t^*,\dots,t^{**}-1\}$. Moreover, by the definition of  stopping time $t^{**}$, $D_t\leq D_{t-L'}/2$ for 
$t\in\{t^*+L',\dots, t^{**}-1\}$, where $L'=O(d \log k)$ (see stopping rule (C)). Hence, $f_t(i,\eta)$ is a non-decreasing sequence, and  $f_t(i,\eta)\geq 4f_{t-L'}(i,\eta)$ for $t\in\{t^*+L',\cdots,t^{**}-1\}$. Let $t'$ be the first step $t$ in $[t^*,t^{**}-1]$ when $f_{t'}(i,\eta) \geq 1$. If $f_t(i,\eta) < 1$ for all $t\in\{t^*,\cdots,t^{**}-1\}$, then $t' = t^{**}$. We have
$$\frac{1}{8}\sum_{t=t^*}^{t^{**}-1}
\frac{\min\{2D_t^2,A_k M_t^2(i,\eta)\}}{ R_t^2} \leq
\sum_{t=t^*}^{t^{**}-1}
\min\{1,f_t(i,\eta)\} = 
\underbrace{\sum_{t=t^*}^{t'-1}
f_t(i,\eta)}_{\Sigma_I}+
\underbrace{\sum_{t=t'}^{t^{**}-1}1}_{\Sigma_{II}}.$$
The first sum ($\Sigma_I$) on the right hand side is upper bounded by $2L'\cdot f_{t'}(i,\eta)$, because $f_t(i,\eta)\geq 4f_{t-L'}(i,\eta)$ for $t < t^{**}$. In turn, $2L'\cdot f_{t'}(i,\eta)\leq 2L'=O(d\log k)$, because $f_t(i,\eta)\leq 1$ for $t<  t'$. The second sum ($\Sigma_{II}$) equals $t^{**} - t'$. Since $f_t(i,\eta) \geq 4f_{t-L'}(i,\eta)$ for every $t\in[t^*+L',t^{**}-1]$, we have
$$
\left\lfloor\frac{(t^{**}-1) - t'}{L'}\right\rfloor\leq 
\log_4 \frac{f_{t^{**}-1}(i,\eta)}{f_{t'}(i,\eta)}\leq 
\log_4 f_{t^{**}-1}(i,\eta) = 
\log_4 \bigg(\frac{A_k\,M^2_{t^{**}-1}(i,\eta)}{D_{t^{**}-1}^2}\bigg)
.
$$
It remains to show that $M_{t^{**}-1}(i,\eta) = O(D_{t^{**}-1})$ and thus 
$$t^{**} - t' = O(L' \log A_k) = O(d\log k \log A_k).$$
We have,
$M_{t^{**}-1}(i,\eta) \leq \|x-c^*\|_2 + D_t \leq 2D_t$,
where we used that for every $t<t^{**}$, $D_t >\|x-c^*\|_2$ (see stopping rule~(C)). This finishes the proof of Lemma~\ref{lem:eq:sum-less-d-logk-loglogk}.
\end{proof}

\subsection{Proof of Lemma~\ref{lem:subst-vars}}\label{sec:lem:subst-vars}
We first make the following simple but crucial observation.
\begin{claim}\label{cl:bound-eta}
If $\widetilde Z_t(i,\sigma,\theta) > 0$, then for $\eta = m_i^t + (\sigma + \varepsilon) \sqrt{\theta}R_t$, we have
$$|\eta - m^t_i|\equiv |(\sigma + \varepsilon) \sqrt{\theta}R_t| \leq \frac{c^*_i - x_i}{\varepsilon}.$$
\end{claim}
\begin{proof}[Proof of Claim~\ref{cl:bound-eta}]
If $\widetilde Z_t(i,\sigma,\theta) > 0$, then the cut with parameters $i$, $\sigma$, $\theta$ separates $x$ and $c^*$ (otherwise, $Z_t(i,\sigma,\theta)$ and $\widetilde Z_t(i,\sigma,\theta)$ would be equal to $0$). That is, 
$x_i \leq m_i^t + \sigma\sqrt{\theta}R_t$
and 
$c^*_i > m_i^t + (\sigma+\varepsilon)\sqrt{\theta}R_t$.
Write,
$$
c^*_i - x_i = (c^*_i - m^t_i) - (x_i - m^t_i) > 
(\sigma + \varepsilon) \sqrt{\theta}R_t - 
\sigma \sqrt{\theta}R_t
= \varepsilon \sqrt{\theta}R_t.
$$
Hence, 
$$|(\sigma + \varepsilon) \sqrt{\theta}R_t|
= \frac{|\sigma + \varepsilon|}{\varepsilon}\cdot \varepsilon \sqrt{\theta}R_t < \frac{|\sigma + \varepsilon|}{\varepsilon} (c_i^* - x_i).$$
\end{proof}
\begin{proof}[Proof of Lemma~\ref{lem:subst-vars}]
We have
$$
\int_0^1 \frac{\widetilde Z_t(i,-1,\theta)
+ \widetilde Z_t(i,1,\theta)}{2}\;d\theta
= \frac{1}{2}\sum_{\sigma \in \{\pm 1\}}
\int_0^1 \widetilde Z_t(i,\sigma,\theta)\;d\theta.
$$
Make the substitutions $\eta_{\sigma} = m^t_i +(\sigma+\varepsilon)R_t\sqrt{\theta}$.
Then, 
$d\theta = \frac{2(\eta_{\sigma}- m_i^t)}{(\sigma+\varepsilon)^2 R_t^2}d\eta_\sigma$ and
$$
\int_0^1 \frac{\widetilde Z_t(i,-1,\theta)
+ \widetilde Z_t(i,1,\theta)}{2}\;d\theta
= \sum_{\sigma \in \{\pm 1\}}
\int_{m_i^t}^{m_i^t+(\sigma+\varepsilon)R_t} \frac{\widetilde Z_t(i,\sigma,\theta)}{(\sigma+\varepsilon)^2 R_t^2}\cdot (\eta_{\sigma}- m_i^t)\;d\eta_{\sigma}.
$$
By Claim~\ref{lem:subst-vars}, $|\eta_{\sigma} - m_i^t|\leq |\sigma  + \varepsilon|/\varepsilon\cdot (c^*_i- x_i)$. Since $Z(i,\sigma,\theta)\geq 0$, we have $\widetilde Z(i,\sigma,\theta) = \max\{Z_t(i_t,\sigma_t,\theta_t) - 4\|x-c^*\|_2^2,0\} \leq Z(i,\sigma,\theta)$. As we discuss in the previous section, $\widetilde Z(i,\sigma,\theta) \leq Z(i,\sigma,\theta) \leq  \min\{2D_t^2,A_kM_t^2(i,\eta_{\sigma})\}$ (see Claim~\ref{cl:two-bounds-on-Z}). Also, if $\eta_{\sigma}\notin [x_i,c_i^*]$, then $x$ and $c^*$ are not separated by the tuple $(i,\sigma, \theta)$, which implies 
$\widetilde Z(i,\sigma,\theta) = 0$.
Thus,
$$
\int_0^1 \frac{\widetilde Z_t(i,-1,\theta)
+ \widetilde Z_t(i,1,\theta)}{2}\;d\theta
\leq \frac{c_i^*-x_i}{\varepsilon(1-\varepsilon)}\; \int_{x_i}^{c^*_i} \frac{\min\{2D_t^2,A_kM_t^2(i,\eta)\}}{ R_t^2} \;d\eta.
$$
This concludes the proof of Lemma~\ref{lem:subst-vars}.
\end{proof}

%% file: paper/lower-bound.tex


\section{Lower Bound on the Bi-criteria Approximation}

In this section, we prove Theorem~\ref{thm:lb_bi_criteria}. We show a lower bound on the price of explainability for $k$-means in the bi-criteria setting. Our proof follows the general approach by~\cite{MS21}.

\medskip

\noindent \textbf{Theorem~\ref{thm:lb_bi_criteria}.}\emph{
For every $k > 500$ and $\ln^3 k / \sqrt{k} < \delta < 1 / 100$, there exists an instance $X$ with $k$ clusters such that the $k$-means cost for every threshold tree $\calT$ with $(1 + \delta)k$ leaves is at least
$$
\mathrm{cost}(X,\calT) \geq \Omega\left(\frac{\log^2 k}{\delta}\right)\OPT_k(X).
$$}
\begin{proof}[Proof of Theorem~\ref{thm:lb_bi_criteria}]
We construct a hard instance for explainable clustering as follows. Let $d = 300 \ceil{\ln k}$. 
Consider the grid $\{0,\varepsilon,2\varepsilon,\dots,1\}^d$ with step size $\varepsilon = 50\delta/\ceil{\ln k}$ in the $d$-dimensional unit cube $[0,1]^d$. We uniformly sample $k$ centers $C = \{c^1,c^2,\dots, c^k\}$ from the nodes of the grid. Then, we create a data set $X$.
For every center $c^i$ in $C$, data set $X$ contains many (namely, $k^2\ceil{\ln^3 k}$) points co-located with $c^i$ and two special points
$c^i\pm (\varepsilon,\varepsilon,\dots,\varepsilon)$. Hence, the total number of points in $X$ is $k^3\ceil{\ln^3 k} + 2k$. Note that all centers and all points in $X$ lie in the nodes of the  grid.

The cost of the $k$-means clustering with centers $C= \{c^1,c^2,\dots, c^k\}$ equals $2kd\varepsilon^2$, since  the distance from the special points $c^i\pm (\varepsilon,\varepsilon,\dots,\varepsilon)$ to $c^i$ is $\varepsilon\sqrt{d}$. Hence, the cost of the optimal $k$-means 
clustering is at most $2kd\varepsilon^2$. We now show that there exists an instance such that the cost of every \emph{explainable} $k$-means clustering with 
$(1+\delta)k$ centers is at least 
$2kd\varepsilon^2 \cdot \Omega(\nicefrac{1}{\delta}\log^2 k)
$.
In this instance, every explainable $k$-means clustering with $(1+\delta)k$ centers separates at least $\delta k = \Omega(\varepsilon k\ln k)$ special points $c^i\pm (\varepsilon,\varepsilon,\dots,\varepsilon)$ from $c^i$. The cost of each special point separated from its original center is at least $\Omega(d)$. Thus, the total cost of every explainable $k$-means clustering is at least $\Omega(d\varepsilon k \ln k) = 2kd\varepsilon^2 \cdot \Omega(\nicefrac{1}{\delta}\log^2 k)$. 
First, we prove that with high probability every two centers in $C$  are far apart.

\begin{lemma}\label{lem:separation_kmeans}
With probability at least $1-1/k^2$ the following statement holds: The distance between every two distinct
centers $c'$ and $c''$ in $C$ is at least $\sqrt{d}/5$.
\end{lemma}

\begin{proof}
We can select a random center in the grid $\{0,\varepsilon,2\varepsilon,\dots,1\}^d$ using the following procedure: First, pick a candidate center uniformly from the cube $[-\varepsilon/2,1+\varepsilon/2]^d$ and then move the chosen point to the closest grid point. Note that the $\ell_2$-distance from every point in this cube to the closest grid point is at most $\sqrt{d}\, \varepsilon/2 \leq  \sqrt{d}/36$ since $\varepsilon \leq 1/18$. 

Consider two distinct centers $c',c'' \in C$. Let $c^*$ and $c^{**}$ be the candidate centers corresponding to $c'$ and $c''$. 
If $\|c^* - c^{**}\|_2 \geq \sqrt{d/12}$, then by the triangle inequality, we have
$$
\norm{c'-c''}_2 \geq \norm{c^*-c^{**}}_2 - \norm{c'-c^*}_2 - \norm{c''-c^{**}}_2 \geq \frac{\sqrt{d}}{\sqrt{12}} - \frac{\sqrt{d}}{18} \geq \frac{\sqrt{d}}{5}.
$$

Thus, we need to show that with probability at least $1-1/k^2$, the $\ell_2$-distance between every two candidate centers uniformly sampled from the cube $[-\varepsilon/2,1+\varepsilon/2]^d$ is at least $\sqrt{d/12}$. Consider two candidate centers $c^*,c^{**}$. Since $c^*,c^{**}$ are chosen uniformly from $[-\varepsilon/2,1+\varepsilon/2]^d$, each coordinate of $c^*,c^{**}$ is drawn from $[-\varepsilon/2,1+\varepsilon/2]$. Hence, we have
$$
\E_{c^*,c^{**}} [\norm{c^*-c^{**}}_2^2] = \sum_{i=1}^d \E_{c^*_i,c^{**}_i} [(c^*_i-c^{**}_i)^2] = d \cdot\frac{(1+\varepsilon)^2}{6}.
$$
Let $X_i = (c^*_i-c^{**}_i)^2/(1+\varepsilon)^2$ for $i \in \{1,\dots,d\}$. Random variables $\{X_i\}_{i=1}^d$ are independent and each $X_i$ lies in $[0,1]$. Thus, by Hoeffding's inequality, we have
$$
\prob{\sum_{i=1}^d X_i - \E\bigg[\sum_{i=1}^d X_i\bigg]  \leq -\sqrt{2d\ln k}} \leq  e^{-4\ln k} = \frac{1}{k^4}.
$$
Since $d = 300\ceil{\ln k}$, the squared distance between $c^*$ and $c^{**}$ is less than $d/12$ with probability at most $1/k^4$.
Using the union bound over all pairs of candidate centers, we conclude that the squared distance between every two candidate centers is at least $d/12$ with probability at least $1-1/k^2$.
\end{proof}

All data points in $X$ are in the grid $\{-\varepsilon, 0, \varepsilon, 2\varepsilon,\dots, 1,1+\varepsilon\}^d$. Every internal node $u$ in the threshold tree should contain a threshold cut that separates at least two data points in that node $u$. Otherwise, we can ignore this threshold cut since one side of this cut contains no data points. If two threshold cuts have the same coordinate and thresholds within the same grid interval $(j\varepsilon,j\varepsilon+\varepsilon)$, then these two threshold cuts create the same partition of data points contained in the internal node. Since there are at most $1/\varepsilon + 2$ different grid intervals for each coordinate, the number of distinct threshold cuts for each internal node is at most $d(1/\varepsilon+2) \leq 2d/\varepsilon$. Every node in the threshold tree corresponds to a cell in $\bbR^d$. This cell is determined by the threshold cuts on the path from the root to that node. Let $\pi$ be an ordered set of tuples $(i_j,\xi_j,\lambda_j)$, where $(i_j,\xi_j)$ is the $j$-th threshold cut on the path from the root to the node, and $\lambda_j \in \{\pm 1\}$ specifies one of the sides of the cut. 
Then, every ordered set $\pi$ corresponds to a path in the threshold tree starting in the root.

Let $u(\pi)$ be the intersection of the cuts in $\pi$. We say that a center $c^i$ in $u(\pi)$ is damaged if one of the special points 
$c^i\pm (\varepsilon,\dots, \varepsilon)$ is separated from $c^i$ by one of the threshold cuts in $\pi$. In other words, 
$c^i$ is damaged if $c^i\in u(\pi)$, but $c^i - (\varepsilon,\dots, \varepsilon)\notin u(\pi)$ or 
$c^i + (\varepsilon,\dots, \varepsilon)\notin u(\pi)$. Otherwise, we say that $c^i$ is not damaged. Similarly, we say that a node of the grid $x\in u(\pi)$ is not damaged if $x \pm (\varepsilon,\dots, \varepsilon)\in u(\pi)$. Let $F_{u(\pi)}$ be the set of all centers that are not damaged in node $u(\pi)$.
We show that with high probability, if a node $u(\pi)$ contains more than $\sqrt{k}$ centers, every threshold cut that splits node $u(\pi)$ damages at least $\varepsilon|F_{u(\pi)}|/2$ centers in $F_{u(\pi)}$. 

\begin{lemma}\label{lem:damage_kmeans}
With probability at least $1-1/k$, the following holds: For every path (ordered set of cuts) $\pi$ of length at most $\log_2 k/4$, we have (a)  $|F_{u(\pi)}|\leq \sqrt{k}$; or (b) every threshold cut that separates at least two data points in $u(\pi)$ damages at least $\varepsilon|F_{u(\pi)}|/2$ centers in $F_{u(\pi)}$.
\end{lemma}
\begin{proof}
Consider a fixed ordered set of cuts $\pi$ of size at most $\log_2 k/4$.
We upper bound the probability that both events (a) and (b) do not occur for this fixed path $\pi$ on the random instance $X$.  If $|F_{u(\pi)}| \leq \sqrt{k}$, then the event (a) happens. So, we assume that $F_{u(\pi)}$ contains more than $\sqrt{k}$ centers. 
We then bound the probability that event (b) happens conditioned on the size of $F_{u(\pi)}$. Observe that all centers in $F_{u(\pi)}$ are distributed uniformly and independently among the grid nodes in $u(\pi)$ that are not damaged by the cuts in $\pi$ conditioned on $|F_{u(\pi)}|$.
Pick an arbitrary threshold cut $(i,\xi)$ in $u(\pi)$ that separates at least two nodes of the grid in $u(\pi)$. For every center $c$ in $F_{u(\pi)}$, the probability that the threshold cut $(i,\xi)$ damages this center $c$ is at least $\varepsilon$. Let $X_j$ be the indicator random variable that the $j$-th center in $F_{u(\pi)}$ is damaged by $(i,\xi)$. The expected number of centers in $F_{u(\pi)}$ damaged by cut $(i,\xi)$ conditioned on $|F_{u(\pi)}| = l$ equals
$$
\E\bigg[\sum_{j=1}^l X_j \Bigm\vert \abs{F_{u(\pi)}} = l \bigg] \geq \varepsilon l.
$$
Let $\mu = \E[\sum_{j} X_j \mid |F_{u(\pi)}| = l]$. By the Chernoff bound for Bernoulli random variables, we have
$$
    \pr\bigg\{\sum_{j=1}^l X_j \leq \varepsilon\abs{F_{u(\pi)}}/2 \Bigm\vert \abs{F_{u(\pi)}} = l \bigg\} \leq \pr\bigg\{\sum_{j=1}^l X_j \leq \mu/2 \Bigm\vert \abs{F_{u(\pi)}} = l \bigg\}
    \leq e^{-\mu/8} \leq e^{-\varepsilon\sqrt{k}/8}.
$$
Combining all conditional probabilities for $|F_{u(\pi)}| > \sqrt{k}$, the probability that the event (b) doesn't happen is at most $e^{-\varepsilon\sqrt{k}/8}$.
Since all data points are in the grid $\{-\varepsilon, 0,\varepsilon,2\varepsilon,\dots,1, 1+\varepsilon\}^d$, there are at most $2d/\varepsilon$ different threshold cuts that separates at least two data points 
in node $u(\pi)$. By the union bound, the  probability that both events (a) and (b) do not happen is at most $e^{-\varepsilon\sqrt{k}/8}\cdot 2d/\varepsilon \leq e^{-2\ln^2 k}$. 
Since there are at most $4d/\varepsilon$ different choices for each tuple $(i_j,\xi_j,\lambda_j)$ in $\pi$, the number of paths with length less than $m = \log_2 k/4$ is at most $m(4d/\varepsilon)^m \leq e^{\ln^2 k}$. Thus, by the union bound over all paths with length less than $\log_2 k /4$, we get that (a) or (b) holds with probability at least
$$
1-m(4d/\varepsilon)^m  \cdot e^{-\varepsilon\sqrt{k}/8}\cdot 2d/\varepsilon  \geq 1 - e^{\ln^2 k} \cdot e^{-2\ln^2 k} \geq 1-\frac{1}{k}.
$$
since $d/\varepsilon \leq 15000\sqrt{k}\ln^3 k$ for $d = 300 \ceil{\ln k}$ and $\varepsilon = 50\delta/\ceil{\ln k} \geq 50\sqrt{k}\ln^2 k$.
\end{proof}

By Lemma~\ref{lem:separation_kmeans} and Lemma~\ref{lem:damage_kmeans}, we can find an instance $X$ such that the following  conditions hold:
\begin{itemize}
    \item The distance between every two distinct centers $c'$ and $c''$ in $C$ is at least $\sqrt{d}/5$.
    \item For every path (ordered set of cuts) $\pi$ of length at most $\log_2 k/4$, we have (a) $|F_{u(\pi)}|\leq \sqrt{k}$; or (b) every threshold cut that separates at least two data points in $u(\pi)$ damages at least $\varepsilon|F_{u(\pi)}|/2$ centers in $F_{u(\pi)}$.
\end{itemize}

We first show that the threshold tree must separate all centers. Suppose there is a leaf contains more than one center. Since the distance between every two centers is at least $\sqrt{d}/5$, there exists at least one center in this leaf with distance greater than $\sqrt{d}/10$ to the optimal center of this leaf. Since we add $k^2\ceil{\ln^3 k}$ points co-located with each center, the cost for the leaf that contains more than one center is greater than $k^2\ceil{\ln^3 k} \cdot d/100 = 2kd\varepsilon^2 \cdot \Omega(\nicefrac{1}{\delta}\log^2 k)$. Thus, the lower bound holds for any threshold tree that does not separate all centers. To separate all centers, the depth of the threshold tree must be at least $\ceil{\log_2 k}$. We show the following lower bound on the number of damaged centers for every threshold tree that separates all centers.

\begin{lemma}\label{lem:damaged_centers}
Consider any instance $X$ with $k$ centers satisfies two conditions in Lemma~\ref{lem:separation_kmeans} and Lemma~\ref{lem:damage_kmeans}. For every threshold tree that separates all centers in $C$, there are at least $2\delta k$ damaged centers.
\end{lemma}
\begin{proof}
Consider any threshold tree $\calT$ that separates all centers. We consider the following two cases. If the number of damaged centers at level $\floor{\log_2 k}/4$ of threshold tree $\calT$ is more than $k/2$, then the total number of damaged centers generated by this threshold tree is more than $2\delta k$.
 
If the number of damaged centers at level $\floor{\log_2 k}/4$ of threshold tree $\calT$ is less than $k/2$, then the number of centers that are not damaged at each level $i= 1,2,\dots,\floor{\log_2 k}/4$ is at least $k/2$. We call a node $u$ a small node if it contains at most $\sqrt{k}$ centers which are not damaged, otherwise we call it a large node. We now lower bound the number of centers damaged at a fixed level $i \in \{1,2,\cdots,\floor{\log_2 k}/4\}$. 
For every level $i\in \{1,2,\cdots,\floor{\log_2 k}/4\}$, the number of nodes at level $i$ is at most $k^{1/4}$.
Since each small node contains at most $\sqrt{k}$ centers that are not damaged, the total number of centers that are not damaged in small nodes at level $i$ is at most $k^{3/4}$. Since the total number of centers that are not damaged at level $i$ is at least $k/2$, the number of centers that are not damaged in large nodes at level $i$ is at least $k/4$. By Lemma~\ref{lem:damage_kmeans}, the number of damaged centers generated at level $i$ is at least $\varepsilon k/8$. Therefore, the total number of damaged centers generated by this threshold tree $\calT$ is at least
$$
\frac{\floor{\log_2 k}}{4}\cdot \frac{\varepsilon k}{8} \geq \frac{50\floor{\log_2 k} \delta k}{32 \ln k} \geq 2\delta k,
$$
which completes the proof.
\end{proof}

We now lower bound the cost for every threshold tree with $(1+\delta)k$ leaves that separates all centers. Consider any threshold tree $\calT$ with $(1+\delta)k$ leaves that separates all centers in $C$. By Lemma~\ref{lem:damaged_centers}, we have more than $2\delta k$ data points separated from their original centers by $\calT$. For each point $x$ separated from its original center $c$, one and only one of the following may occur: (1) the data point $x$ is assigned to a leaf containing a center $c' \neq c$; (2) the data point $x$ is assigned to a leaf containing no center. Among these $2\delta k$ data points, we show that there are at least $\delta k$ data points that have distances to their new centers greater than $\sqrt{d}/20$.  

For each leaf containing a center $c'$, the optimal center for this leaf is shifted from $c'$ by at most $\varepsilon\sqrt{d}$. Otherwise, the cost of this leaf is at least $k^2\ceil{\ln^3 k} \cdot\varepsilon^2 d = 2kd\varepsilon^2 \cdot \Omega(\nicefrac{1}{\delta}\log^2 k)$ since there are $k^2\ceil{\ln^3 k}$ data points co-located at each center. Suppose a point $x$ separated from its original center $c$ is assigned to a leaf containing a center $c' \neq c$. By Lemma~\ref{lem:separation_kmeans} and the triangle inequality, the distance from the point $x$ to the optimal center for this leaf is at least $\sqrt{d}/10$. 

For each leaf containing no center, it may contain several points from distinct clusters. Among these points, there is at most one point within $\sqrt{d}/20$ distance  of the optimal center for this leaf. Suppose two points $x'$ and $x''$ from distinct clusters are within $\sqrt{d}/20$ distance of the optimal center for this leaf. Then, the distance between $x'$ and $x''$ is at most $\sqrt{d}/10$. Let $c'$ and $c''$ be the original centers for points $x'$ and $x''$ respectively. The distance between $c'$ and $c''$ is at most $\sqrt{d}/10 + 2\varepsilon\sqrt{d} \leq \sqrt{d}/5$, which contradicts the distance between every two centers is at least $\sqrt{d}/5$.

Since the threshold tree $\calT$ has $(1+\delta)k$ leaves, there are $\delta k$ leaves that do not contain a center. Thus, among points separated from their original centers, there are at most $\delta k$ points with distance less than $\sqrt{d}/20$ to their new centers. Since there are more than $2\delta k$ points separated from their original centers, we have at least $\delta k$ points with cost greater than $d/400$. Therefore, the cost given by this threshold tree $\calT$ is at least
$$
\mathrm{cost}(X,\calT) \geq \delta k \cdot \frac{d}{400} = \Omega(\delta d k).
$$
Recall that the optimal $k$-means cost for this instance is at most $2k\varepsilon^2 d$ and $\varepsilon = 50\delta/\ceil{\ln k}$. Thus, the cost given of this explainable clustering is at least
$$
\mathrm{cost}(X,\calT) = \Omega(\delta d k) \geq \Omega\left( \frac{\log^2 k}{\delta}\right) {\OPT}_k(X).
$$
\end{proof}

%% file: paper/lb-ExKMC.tex
\section{Lower Bound for the ExKMC Algorithm}\label{sec:greedy-lb}
In this section, we show the lower bound for the ExKMC algorithm. The ExKMC algorithm is an expanding explainable $k$-means algorithm proposed by \citet{frost2020exkmc}. Given a parameter $k'>k$ as the number of leaves, the ExKMC outputs a threshold tree $T$ with $k'$ leaves. 
We consider the ExKMC algorithm that starts from the base tree given by the IMM algorithm in~\citet{DFMR20}. The IMM algorithm iteratively chooses the threshold cut that minimizes the number of mistakes, where a mistake means a point is separated from its original center. For any threshold tree with more than $k$ leaves, the ExKMC algorithm considers the surrogate cost, which is the cost by assigning each leaf to its best center in $C$. Then, the ExKMC algorithm iteratively chooses the threshold cut that minimizes the surrogate cost.
Our proof is inspired by the constructions in~\citet{EMN22},~\citet{laber2021price} and~\citet{CharikarHu21}.

\begin{theorem}
For every $k>10$, and $\delta\in (0,1/4)$, there exists an instance $X$ with $k$ clusters such that the $k$-means cost for the threshold tree $\calT$ returned by the ExKMC algorithm with an IMM base tree and $k' = (1+\delta) k$ leaves is at least
$$
\mathrm{cost}(X,\calT) \geq \Omega\left((1-4\delta)\cdot\frac{k^2}{\log k}\right)\OPT_k(X).
$$
\end{theorem}

\noindent\textbf{Remark:} This provides a $\tilde{\Omega}(k^{2})$ lower bound for the ExKMC algorithm when $\delta \in (0,1)$ is a constant and $k \to \infty$.


\begin{proof}
We first construct $k$ centers for the instance. Without loss of generality, we assume $k = 2\tilde{k} + 1$ is an odd number. Let $p = 3\log_2 k$ and $d = \tilde{k}+p-1$. Then, we choose $k$ centers $C = \{c^1,c^2,\cdots, c^k\}$ in the $d$-dimensional space $\bbR^d$. Let the first $\tilde{k}-1$ coordinates of $c^1$ be all zeros $(0,0,\cdots,0)$. For each $i \in \{2,\cdots, \tilde{k}+1\}$, let the first $\tilde{k}-1$ coordinates of center $c^i$ be the same as those of $e_{i-1}$ the identity vector on the $(i-1)$-th coordinate. For every coordinate $j \in \{\tilde{k},\tilde{k}+1,\cdots,\tilde{k}+p-1\}$, we pick a random permutation $\sigma$ of $\{0,1,\cdots,\tilde{k}\}$ and assign the $j$-th coordinate of centers $c^1,c^2,\dots,c^{\tilde{k}+1}$ be this random permutation, i.e. $c^i_j = \sigma(j)$. For each $i \in \{\tilde{k}+2,\cdots, 2\tilde{k}+1\}$, the first $\tilde{k}-1$ coordinates of $c^i$ are all zero, and the rest $p$ coordinates of center $c^i$ are identical to those of the center $c^{i-\tilde{k}}$.

We now construct the instance $X$ as follows. For the center $c^1$ and every coordinate $j \in \{1,2,\cdots, \tilde{k}-1\}$, we add one data point at $e_j$. For every center $c^i$, and every coordinate $j \in \{\tilde{k},\tilde{k}+1,\cdots, \tilde{k}+p-1\}$, we add two data points at $c^i+e_j$ and two data points at $c^i-e_j$. For every center $c^i$, we also add many data points co-located with $c^i$.  

For this instance $X$, the cost of the $k$-means clustering with centers $C = \{c^1,c^2,\cdots, c^k\}$ equals $(\tilde{k}-1)+4p\tilde{k}$. Thus, the optimal $k$-means cost of $X$ is at most $(\tilde{k}-1)+4p\tilde{k} = O(\tilde{k}\log k)$. Let $\calT$ be the threshold tree returned by the ExKMC algorithm with the IMM base tree and $k' = (1+\delta)k$ leaves. We show that the cost of the threshold tree $\calT$ is at least $\Omega((1-\delta)k^3)$. We first show that with high probability every two centers in $\{c^1,c^2,\cdots,c^{\tilde{k}+1}\}$ are far apart.

\begin{lemma}\label{lem:exkmc_separation}
With probability at least $1-1/k$ the following statement holds: The distance between every two distinct
centers $c'$ and $c''$ in $\{c^1,c^2,\cdots,c^{\tilde{k}+1}\}$ is at least $k/5$.
\end{lemma}

\begin{proof}
Consider two distinct centers $c',c''$ in $\{c^1,c^2,\cdots,c^{\tilde{k}+1}\}$. For every coordinate $j \in \{\tilde{k},\tilde{k}+1,\cdots, \tilde{k}+p-1\}$, the $j$-th coordinate of centers $\{c^1,c^2,\cdots,c^{\tilde{k}+1}\}$ form a random permutation of $\{0,1,\cdots, \tilde{k}\}$. Thus, we have for every $j \in \{\tilde{k},\tilde{k}+1,\cdots, \tilde{k}+p-1\}$
$$
\prob{|c'_j-c''_j| \geq \frac{\tilde{k}}{2}} = \frac{1}{2}.
$$
The distance between $c'$ and $c''$ is at least $\nicefrac{\tilde{k}}{2}$ with probability $1-(\nicefrac{1}{2})^p = 1-\nicefrac{1}{k^3}$. By the union bound over all pairs of centers in $\{c^1,c^2,\cdots,c^{\tilde{k}+1}\}$, the distance between two distinct centers in $\{c^1,c^2,\cdots,c^{\tilde{k}+1}\}$ is at least $\nicefrac{k}{5}$ with probability at least $1-\nicefrac{1}{k}$. 
\end{proof}


By Lemma~\ref{lem:exkmc_separation}, we can find an instance $X$ such that the distance between every two distinct centers $c'$ and $c''$ in $\{c^1,c^2,\cdots,c^{\tilde{k}+1}\}$ is at least $k/5$. Then, we show that there are at least $(1-4\delta)\tilde{k}$ data points which are separated from their original centers in the threshold tree $\calT$ given by the ExKMC algorithm with the IMM base tree. The algorithm first uses the IMM algorithm in~\citet{DFMR20} to generate a threshold tree with $k$ leaves. The IMM algorithm iteratively chooses the threshold cut that minimizes the number of mistakes to separate centers, where a mistake means a data point is separated from its original center. 

For this instance $X$, we show that the first $\tilde{k}-1$ cuts chosen by the IMM algorithm are at the first $\tilde{k}-1$ coordinates. At any iteration $t \leq \tilde{k}-1$, suppose the first $t-1$ cuts are at the first $\tilde{k}-1$ coordinates. If any center $c^i$ for $i \in \{2,\cdots,\tilde{k}\}$ is not separated from center $c^1$, then the threshold cut at coordinate $i-1$ will separate center $c^i$ from other centers and split one data point at $e_j$ from its center $c^1$. Note that centers $c^1$ and $c^{\tilde{k}+2},c^{\tilde{k}+3},\dots,c^{k}$ are not separated at iteration $t$. For every coordinate $j \in \{\tilde{k},\cdots,d\}$, the $j$-th coordinate of these centers form a permutation of $\{0,1,\cdots,\tilde{k}\}$. Therefore, every threshold cut at coordinate $j \in \{\tilde{k},\tilde{k}+1,\cdots, d\}$ will split at least two data points from their centers. Thus, the IMM algorithm will choose a threshold cut at coordinate $i-1 \leq \tilde{k}-1$ at iteration $t$. 

We now bound the number of mistakes in the tree $\calT$ given by the ExKMC algorithm. Since the IMM algorithm chooses the first $\tilde{k}-1$ threshold cuts at the first $\tilde{k}-1$ coordinates, the IMM algorithm splits $\tilde{k}-1$ data points at $e_1, e_2,\dots, e_{\tilde{k}-1}$ from their original center $c^1$.
Since all these $\tilde{k}-1$ data points are separated in $\tilde{k}-1$ leaves of the IMM tree, the ExKMC algorithm with $(1+\delta)k$ leaves can rearrange at most $\delta k$ data points among these $\tilde{k}-1$ data points to their original centers. Therefore, there are at least $\tilde{k}-1 - \delta k \geq (1-4\delta)\tilde{k}$ data points separated from their original centers in the threshold tree $\calT$ given by the ExKMC algorithm with the IMM base tree. 

By Lemma~\ref{lem:exkmc_separation}, the cost of each data point separated from its original center is at least $\Omega(k^2)$. Since $\OPT_k(X) = O(\tilde{k}\log k)$, the cost of the threshold tree $\calT$ is at least 
$$
\cost(X,\calT) \geq \Omega((1-4\delta)\tilde{k} \cdot k^2) \geq \Omega((1-4\delta)k^2/ \log k)\OPT_k(X).
$$
\end{proof}